\documentclass{article}

\PassOptionsToPackage{numbers, compress}{natbib}
 \usepackage[preprint]{neurips_2026}

\usepackage[utf8]{inputenc} 
\usepackage[T1]{fontenc}    
\usepackage{hyperref}       
\usepackage{url}            
\usepackage{booktabs}       
\usepackage{amsfonts}       
\usepackage{nicefrac}       
\usepackage{microtype}      
\usepackage{xcolor}         

\usepackage{graphicx}
\usepackage{subcaption}

\usepackage{amsmath}
\usepackage{amssymb}
\usepackage{mathtools}
\usepackage{amsthm}

\usepackage{algorithm}
\usepackage{alphalph}
\usepackage{algorithmic}
\usepackage{bbding}
\usepackage{makecell}
\usepackage[accsupp]{axessibility}
\usepackage{wrapfig}
\usepackage{orcidlink}

\usepackage{enumitem}

\usepackage{url}            
\usepackage{booktabs}       
\usepackage{amsfonts}       
\usepackage{nicefrac}       
\usepackage{microtype}  

\usepackage{amssymb}
\usepackage{pifont}

\usepackage{graphicx}
\graphicspath{{./images/}}
\usepackage{multirow}

\usepackage[table,dvipsnames]{xcolor}
\definecolor{LightCyan}{rgb}{0.91,0.91,0.98}
\definecolor{LightYellow}{rgb}{1.0, 1.0, 0.88}
\definecolor{magicmint}{rgb}{0.67, 0.94, 0.82}
\definecolor{lightmauve}{rgb}{0.86, 0.82, 1.0}
\definecolor{grannysmithapple}{rgb}{0.66, 0.89, 0.63}
\definecolor{isabelline}{rgb}{0.95,0.93,0.91}

\usepackage[capitalize,noabbrev]{cleveref}

\title{RCoT-Seg: Reinforced Chain-of-Thought for Video Reasoning and Segmentation}

\author{%
  Junwei Wen$^{1}$\thanks{Equal contribution.} \quad
  Deshui Miao$^{1}$\footnotemark[1] \quad
  Guangming Lu$^{1}$ \quad
  Xin Li$^{1,2,3}$\thanks{Corresponding authors.} \quad
  Wenjie Pei$^{1,2}$\footnotemark[2] \\
  $^1$Harbin Institute of Technology, Shenzhen \quad
  $^2$Pengcheng Laboratory \\
  $^3$Pazhou Lab (Huangpu)
}

\begin{document}

\maketitle

\begin{abstract}
Video Reasoning Segmentation (VRS) aims to segment target objects in videos based on implicit instructions that convey human intent and temporal logic.
Existing MLLM-based methods predict masks with a [SEG] token after selecting frames via simple sampling or an auxiliary MLLM, where limited supervision and frame-language similarity rules often yield narrow-scope keyframe choices that weaken holistic temporal understanding and lead to brittle localization in complex multi-object scenes.
To address these issues, we introduce RCoT-Seg, a video-of-thought framework that factorizes VRS into temporal video reasoning (TVR) and keyframe target perception (KTP), explicitly separating temporal reasoning from spatial perception.
Specifically, in the TVR stage, an agentic keyframe selection module, initialized with a curated CoT-start corpus and refined by GRPO under task-aligned rewards, is proposed to generate and reselect the keyframe through self-evaluation, strengthening moment localization and temporal reasoning.
In the KTP stage, RCoT-Seg performs high-resolution segmentation on the selected frame and propagates masks with SAM2-based methods across the sequence, replacing heuristic sampling and external selectors while improving spatial precision and inter-frame consistency.
Extensive experimental results demonstrate that the proposed RCoT-Seg achieves favorable performance against the state-of-the-art methods. 
The code and models will be publicly released at \url{https://github.com/Victor-wjw/RCoT-Seg}.
\end{abstract}

\section{Introduction}

Video reasoning segmentation (VRS)~\cite{yan2024visa, zheng2024villa} aims to generate pixel-level mask sequences from natural-language queries that involve commonsense grounding and implicit temporal logic, typically with multimodal large language models (MLLMs)~\cite{yan2024visa, videolisa}. 
Despite recent progress, VRS remains challenging because the model must identify the temporally valid evidence implied by the query and further ground the target object with fine-grained spatial precision across diverse video scenarios.

Most existing methods~\cite{yan2024visa, VRS-HQ, lin2025glus} follow a ``sample-then-segment'' paradigm, where sparse frames are first obtained by uniform sampling, auxiliary MLLM filtering, or vision-language similarity-based screening, and then used for target localization and mask generation. 
However, such frame selection is usually treated as a one-shot pre-processing step rather than a verifiable decision. Once the selected frame fails to satisfy the language-conditioned temporal requirement, the downstream segmentation stage has limited ability to recover the missing evidence. 

Recently, a concurrent work Veason-R1~\cite{gong2025reinforcing} shows that CoT initialization and GRPO can improve VRS by encouraging reasoning before segmentation. Nevertheless, it still operates on a sparse sampled view and commits to keyframe localization and spatial grounding within a single reasoning trajectory. As a result, keyframe reliability is not explicitly modeled as a controllable intermediate state, making it difficult to diagnose an unreliable keyframe or request an alternative temporal observation.
\begin{figure}
  \centering
  \includegraphics[width=\linewidth]{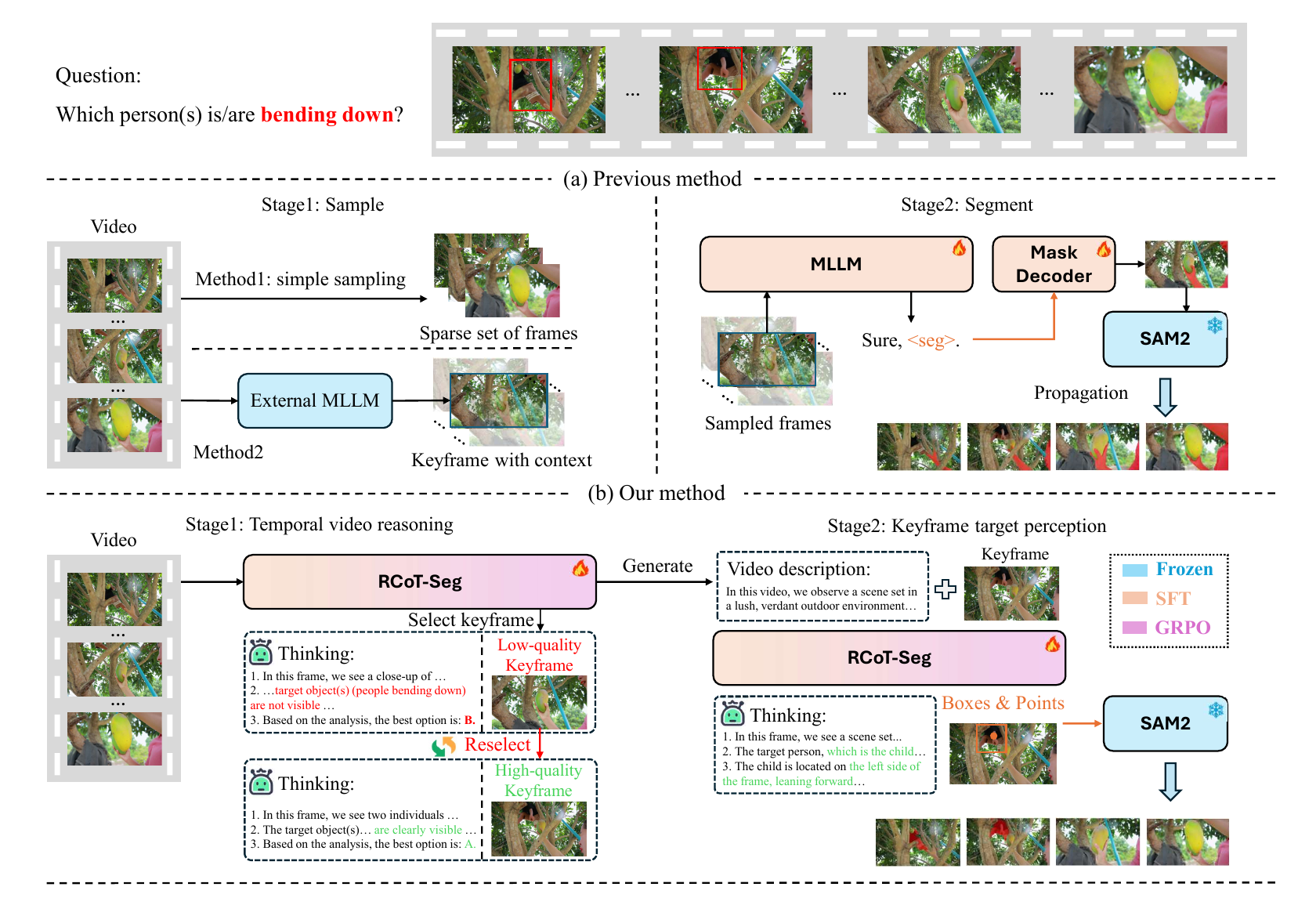}
  \caption{\textbf{Comparison between RCoT-Seg and previous methods.}
  RCoT-Seg segments the correct target through explicit CoT reasoning and concise keyframe selection.}
  \label{fig:framework_comparison}
\end{figure}
To address these limitations, we propose \textbf{RCoT-Seg}, a reinforcement learning framework that treats VRS as an explicit chain-of-thought (CoT) process and decomposes it into two coupled sub-processes: \textit{temporal video reasoning} (TVR) and \textit{keyframe target perception} (KTP).
RCoT-Seg resolves this tension by making CoT an \emph{actionable intermediate state}—a semantic scaffold that is produced by TVR and consumed by KTP.
Instead of relying on pre-sampling, the TVR stage scans the video to infer event structure and temporal conditions implied by the query, and selects a keyframe that best satisfies these conditions.
The KTP stage then operates exclusively on this keyframe with high-resolution perception to produce accurate masks.

A central component of RCoT-Seg is an \textbf{agentic keyframe selection} mechanism with a lightweight self-critical loop.
Beyond outputting a keyframe index, the selector learns to assess whether the current selection is \emph{satisfied} or \emph{unsatisfied} with respect to the query, and triggers \emph{re-selection} when the evidence is insufficient.
This self-evaluation capability turns keyframe selection from a one-shot heuristic into a verifiable decision process, substantially improving reliability under ambiguous temporal language, occlusion, and multi-object distractions.
For segmentation, KTP conditions on both the selected keyframe and a compact video description generated in TVR, using this CoT-derived semantic scaffold to disambiguate targets and guide mask generation while retaining global context.

To enable structured multi-task reasoning, we curate a \textbf{28K-sample CoT dataset} to cold-start a base Qwen2.5-VL-3B model with supervised fine-tuning.
Using customized prompt templates, Qwen2.5-VL-7B generates CoT reasoning traces for both agentic keyframe selection and keyframe target grounding.
We then boost the model with GRPO-based reinforcement learning using task-specific datasets and rewards.
In particular, for keyframe target grounding, we design a Hungarian algorithm-based reward to robustly handle multi-object scenarios and complex visual settings, providing stable credit assignment for accurate matching between predicted masks and target references.
Extensive experiments demonstrate that RCoT-Seg achieves state-of-the-art performance across video reasoning segmentation and referring segmentation benchmarks.

Our contributions are summarized as follows:
\begin{itemize}[leftmargin=*,nosep]
\item We propose \textbf{RCoT-Seg}, a reasoning-enhanced VRS framework that goes beyond single-pass reasoning by factorizing segmentation into \textit{temporal video reasoning} and \textit{keyframe target perception}. This design turns CoT from a passive explanation into an actionable intermediate state that bridges video-level temporal understanding and high-resolution mask generation.

\item We develop \textbf{Agentic Keyframe Selection}, a self-checkable keyframe decision mechanism that explicitly judges whether the selected frame provides sufficient evidence for the query and triggers re-selection when necessary. This improves robustness over one-pass keyframe localization under implicit temporal instructions, occlusion, and multi-object distractions.
\item We introduce a \textbf{matching-aware reward} (Hungarian-based) for RL refinement in keyframe grounding, and show strong empirical gains, achieving state-of-the-art results on multiple VRS and referring segmentation benchmarks.
\end{itemize}

\section{Related work}

\subsection{Video reasoning segmentation}
Video Reasoning Segmentation (VRS) is an emerging task that requires multimodal reasoning to segment target objects in videos based on natural language queries~\cite{videolisa,yan2024visa,zheng2024villa}. Pioneering works like VISA~\cite{yan2024visa} combine an additional MLLM for keyframe sampling, followed by another MLLM for temporal reasoning and segmentation, and employ an object tracker for mask propagation. VideoLISA~\cite{videolisa} introduces a sparse-to-dense sampling strategy and a One-Token-Seg-All paradigm, achieving video-level segmentation through a unified token representation. HyperSeg~\cite{hyperseg} generalizes the unified reasoning framework via a hybrid entity recognition mechanism, supporting cross-domain segmentation and enabling universal segmentation for both image and video inputs. Further advancements include GLUS~\cite{lin2025glus}, which provides MLLM with both global and local contexts that are further enhanced with end-to-end optimized VOS memory modules to improve the consistency, and VRS-HQ~\cite{VRS-HQ}, which enhances temporal consistency through token fusion and utilizes SAM2 for occlusion-aware keyframe selection.

Token-based designs compress object information into a single token, which weakens alignment to the true referent with no explicit chain of thought. In addition,  these methods either sample frames in ways that break temporal coherence or offload keyframe selection to auxiliary MLLMs, undermining end-to-end learning. Veason-R1~\cite{gong2025reinforcing} still treats keyframe localization as a single-pass committed prediction. Once the selected keyframe is unreliable, the subsequent grounding stage has no explicit mechanism to detect the failure or request another temporal observation.Therefore, we design a comprehensive chain-of-thought reasoning pipeline and apply GRPO to improve its robustness, enabling explicit reasoning within a single model.

\subsection{Visual fine-tuning via RL}
Reinforcement learning~\cite{sutton1998reinforcement} has become a significant paradigm for optimizing large language models, particularly demonstrating remarkable effectiveness in enhancing reasoning capabilities, as evidenced by ChatGPT-o1~\cite{jaech2024openai}. Deepseek-R1~\cite{guo2025deepseek} introduces group relative policy optimization (GRPO), which leverages verifiable reward signals to estimate relative advantages among responses, thereby substantially improving reasoning. Building on this, GRPO fine-tuning techniques have been extended to multimodal tasks, covering areas such as image spatial reasoning~\cite{liu2025visual,wang2025pixelthink}, video understanding~\cite{feng2025video,li2025videochat,wang2025time}, multi-image localization~\cite{bai2025univg,zhang2025improving}, and visual generation~\cite{fang2025got,xiao2025mindomni,xue2025dancegrpo}, demonstrating its strong adaptability in complex multimodal scenarios. 
Early work, such as Seg-Zero~\cite{liu2025seg} and VisionReasoner~\cite{liu2025visionreasoner}, designed task-specific rewards for segmentation, improving image-level reasoning and mask quality. Building on this idea, Omni-R1~\cite{zhong2025omni} and Veason-R1~\cite{gong2025reinforcing} adopted GRPO for Ref-AVS and VRS. Omni-R1 follows a cascaded dual-system design that depends on large training corpora and a strong pre-trained VRS backbone. Veason-R1~\cite{gong2025reinforcing} further introduces CoT initialization and GRPO into VRS, demonstrating that reinforcement learning can improve keyframe localization and spatial grounding. Nevertheless, its inference process still produces keyframe localization and object grounding as a single committed reasoning trajectory, without explicitly modeling whether the selected keyframe is reliable enough for downstream segmentation. RCoT-Seg differs by making keyframe reliability an explicit decision state and introducing an agentic re-selection loop, thereby enabling the model to verify and revise temporal evidence before high-resolution target perception. In contrast, RCoT-Seg unifies temporal reasoning and segmentation in a single multi-task model, achieving stronger results in complex video scenarios with only modest training data.
\section{Method}

\begin{figure*}
  \centering
  \resizebox{0.9\linewidth}{!}{
    \includegraphics{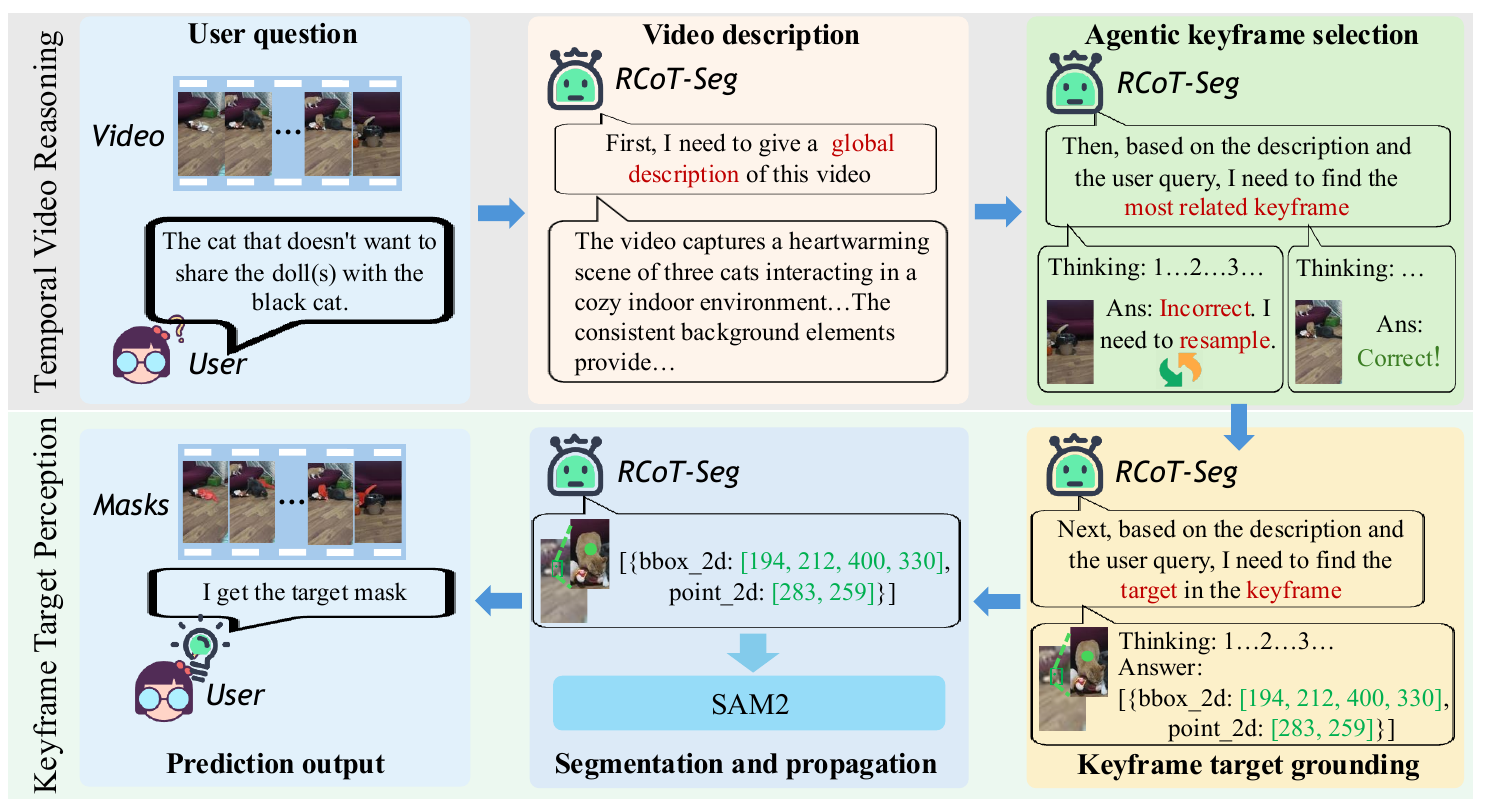}
  }
  \caption{\textbf{Overview of RCoT-Seg unified multi-task pipeline for VRS task.}}
  \label{fig:pipline}
  \vspace{-5mm}
\end{figure*}


Given a video sequence and the query text of the target object(s), the goal of the proposed RCoT-Seg is to precisely localize the target object(s) with masks in each frame. To this end, the proposed RCoT-Seg first generates a video description as a compressed representation of the video content and selects a keyframe via agentic keyframe selection mechanism, then the object information of the target object(s) in the keyframe is obtained, which subsequently serves as prompts for the SAM2~\cite{ravi2024sam} to perform mask generation and propagation, thereby achieving accurate segmentation of the target object(s).

\subsection{Overview}

\cref{fig:pipline} illustrates the unified multi-task pipeline of our RCoT-Seg, consisting of two core stages: \emph{Temporal Video Reasoning} and \emph{Keyframe Target Perception}. 
Given a video sequence $\mathcal{V}\in\mathbb{R}^{T\times3\times H\times W}$ and an instruction text $x_{ins}$, RCoT-Seg first generates the video description $ x_{vid}$, and the candidate keyframe $ z_{sel}$ which is evaluated and refined as final keyframe $ z_{key}$ via agentic keyframe selection mechanism. Then, RCoT-Seg predicts bounding boxes and points $\{(box_i,\ point_i)\}_{i=1}^{N_{\text{obj}}}$ of the target object(s) based on $ z_{key}$ and $ x_{vid}$, which finally serves as prompts for the SAM2 to predict the segmentation masks $\mathcal{M}\in\mathbb{R}^{T\times H\times W}$.

As illustrated in \cref{fig:training_strategy}, we adopt a two-stage training paradigm. We initialize our framework with Qwen2.5-VL-3B~\cite{bai2025qwen2} as the MLLM. In the first stage, we construct a hybrid chain-of-thought dataset and perform supervised fine-tuning. In the second stage, we apply GRPO~\cite{shao2024deepseekmath,guo2025deepseek} to further enhance the capability in judging keyframe selection states and spatiotemporal localization of target objects.
\subsection{Temporal video reasoning} 
\textbf{Initial video reasoning.} RCoT-Seg first exploits the inherent video reasoning capability of the foundation model via carefully designed prompts. Given a low-resolution, unsampled video sequence $\mathcal{V}$, our RCoT-Seg model $\varphi_\theta$ performs video description generation and keyframe generation:
\begin{equation}
    x_{vid}=\varphi_\theta(\mathcal{V},\, p_{vdg}), \qquad 
    z_{sel}=\varphi_\theta(\mathcal{V},\, x_{ins},\, p_{kfg}),
    \label{eq:vid-desc-and-kfg}
\end{equation}
where $p_{vdg}$ and $p_{kfg}$ denote the prompts for video description and keyframe generation, respectively. The description $x_{vid}$ acts as a compact representation of global semantics and the $z_{sel}$ acts as the candidate keyframe.

\noindent\textbf{Agentic keyframe selection (AKS).} 
%
Although $z_{sel}$ is obtained through the above reasoning, we empirically observe that the chosen keyframe can sometimes be suboptimal, and the one-step selection paradigm suffers from a lack of error tolerance.
Therefore, an agentic mechanism is developed, which enables the model to autonomously evaluate and refine the chosen keyframe. 


After the candidate keyframe is selected, RCoT-Seg decides whether the frame belongs to the valid keyframe set or its complement:
\begin{equation}
    x^{A}_{\text{think}},\ \textit{ans}=\varphi_\theta(z_{sel},\, x_{vid},\, x_{ins},\, p_{aks}),\quad \textit{ans}\in\{A,B\},
    \label{eq:agent}
\end{equation}
If \textit{ans} $=A$, the frame is retained for segmentation (see \cref{eq:keyframe-location}); if \textit{ans} $=B$, the model reselects a new keyframe by updating the prompt and invoking
\begin{equation}
    z_{sel}^\prime=\varphi_\theta(\mathcal{V},\, x_{txt},\, p_{kfs}^\prime),
    \label{eq:kf-re-sel}
\end{equation}
and the evaluation in \cref{eq:agent} is repeated until a satisfactory frame is obtained or a maximum iteration count $\lambda$ is reached.

\noindent\textbf{AKS GRPO reward.} We use a two-component reward for AKS under GRPO, a format reward $\mathcal{R}_f$ and an answer-accuracy reward $\mathcal{R}_a$. 
$\mathcal{R}_f$ enforces structured outputs, which means the reasoning must appear inside \texttt{<think>}...\texttt{</think>} and the final decision inside \texttt{<answer>}...\texttt{</answer>}, and the decision must exactly match one of the prompt options. 
$\mathcal{R}_a$ enforces answer correctness as:
\begin{equation}
\mathcal{R}_a=\begin{cases}
1.0, & ans=ans^{GT},\\
0.0, & \text{otherwise.}
\end{cases}
\label{eq:Ra-agent}
\end{equation}
where $ans$ denotes the predicted option and $ans^{GT}$ denotes the ground truth.

\subsection{Keyframe target perception} 
\textbf{Keyframe target grounding (KTG).} RCoT-Seg receives a high-resolution keyframe $z_{key}$, the instruction $x_{ins}$, and the video description $x_{vid}$ as context.  
Given a grounding prompt $p_{ktg}$, it outputs a thinking process and a list of cues:
\begin{equation}
    x^{G}_{\text{think}},\ \{(box_i,\ point_i)\}_{i=1}^{N_{\text{obj}}}
    = \varphi_\theta\!\left(z_{key},\, x_{vid},\, x_{ins},\, p_{ktg}\right),
    \label{eq:keyframe-location}
\end{equation}
where $\{(box_i,\ point_i)\}_{i=1}^{N_{\text{obj}}}$ denotes the predicted bounding boxes and points for $N_{\text{obj}}$ target objects in the keyframe, and $x^{G}_{\text{think}}$ is the generated reasoning trace.

\noindent\textbf{Keyframe segmentation and propagation.} We then apply a frozen SAM2 model to segment each target and propagate masks across the video, using $(box_i, point_i)$ as prompts:
\begin{equation}
    \mathcal{M}_i
    = \varphi_{\text{SAM2}}\!\left(z_{key},\, (box_i,\, point_i),\, \mathcal{V}\right),
    \label{eq:i-th-mask-pre}
\end{equation}
where $\varphi_{\text{SAM2}}$ denotes SAM2 and $\mathcal{V}$ is the video.  
The final prediction is obtained by aggregating per-object masks:
\begin{equation}
    \mathcal{M}=\bigcup_{i=1}^{N_{\text{obj}}}\mathcal{M}_i.
    \label{eq:mask-merge}
\end{equation}

\noindent\textbf{KTG GRPO reward.} To achieve KTG, the reward consists of a format reward $\mathcal{R}_f$ and an answer accuracy reward $\mathcal{R}_a$. $\mathcal{R}_f$ requires the same \texttt{<think>} / \texttt{<answer>} structure. The final answer must be a list of dictionaries with 2D boxes and points, such as
\texttt{[ \{``bbox\_2d'':[x1,y1,x2,y2], ``point\_2d'':[x,y]\}, \dots ]}. $\mathcal{R}_a$ is designed to accommodate multi-object scenarios. 
Let the prediction be $\mathcal{P}_{BP}=\{(box_i,point_i)\}_{i=1}^{N_{\mathrm{obj}}}$ and the ground truth
$\mathcal{P}_{BP}^{GT}=\{(box^{GT}_j,point^{GT}_j)\}_{j=1}^{N_{\mathrm{obj}}^{GT}}$ on the resized $840{\times}840$ scale. Define binary score matrices (1=correct, 0=incorrect):
\begin{gather}
S^{\mathrm{IoU}}_{i,j}=I_{\mathrm{IoU}>\eta}(box_i,box^{GT}_j),\\
S^{\mathrm{boxL1}}_{i,j}=I_{\mathrm{L1}<\gamma_{box}}(box_i,box^{GT}_j),\\[-2pt]
S^{\mathrm{ptL1}}_{i,j}=I_{\mathrm{L1}<\gamma_{pt}\ \wedge\ \mathrm{in\text{-}box}}(point_i,point^{GT}_j).
\end{gather}

Aggregate them with weights $\alpha_{\mathrm{IoU}},\alpha_{\mathrm{boxL1}},\alpha_{\mathrm{ptL1}}$:
\begin{equation}
S_{i,j}=\alpha_{\mathrm{IoU}}S^{\mathrm{IoU}}_{i,j}
+\alpha_{\mathrm{boxL1}}S^{\mathrm{boxL1}}_{i,j}
+\alpha_{\mathrm{ptL1}}S^{\mathrm{ptL1}}_{i,j}.
\end{equation}
We obtain a one-to-one assignment by maximizing $S$ with the Hungarian algorithm, yielding matched pairs $C'$. The answer accuracy reward averages the matched scores:
\begin{equation}
\mathcal{R}_a=\frac{1}{\max\{N_{\mathrm{obj}},\,N_{\mathrm{obj}}^{GT}\}}
\sum_{(i,j)\in C'} S_{i,j}.
\label{eq:Ra-seg}
\end{equation}

\subsection{Training strategy} 

\begin{figure*}
  \centering
  \resizebox{\linewidth}{!}{
    \includegraphics{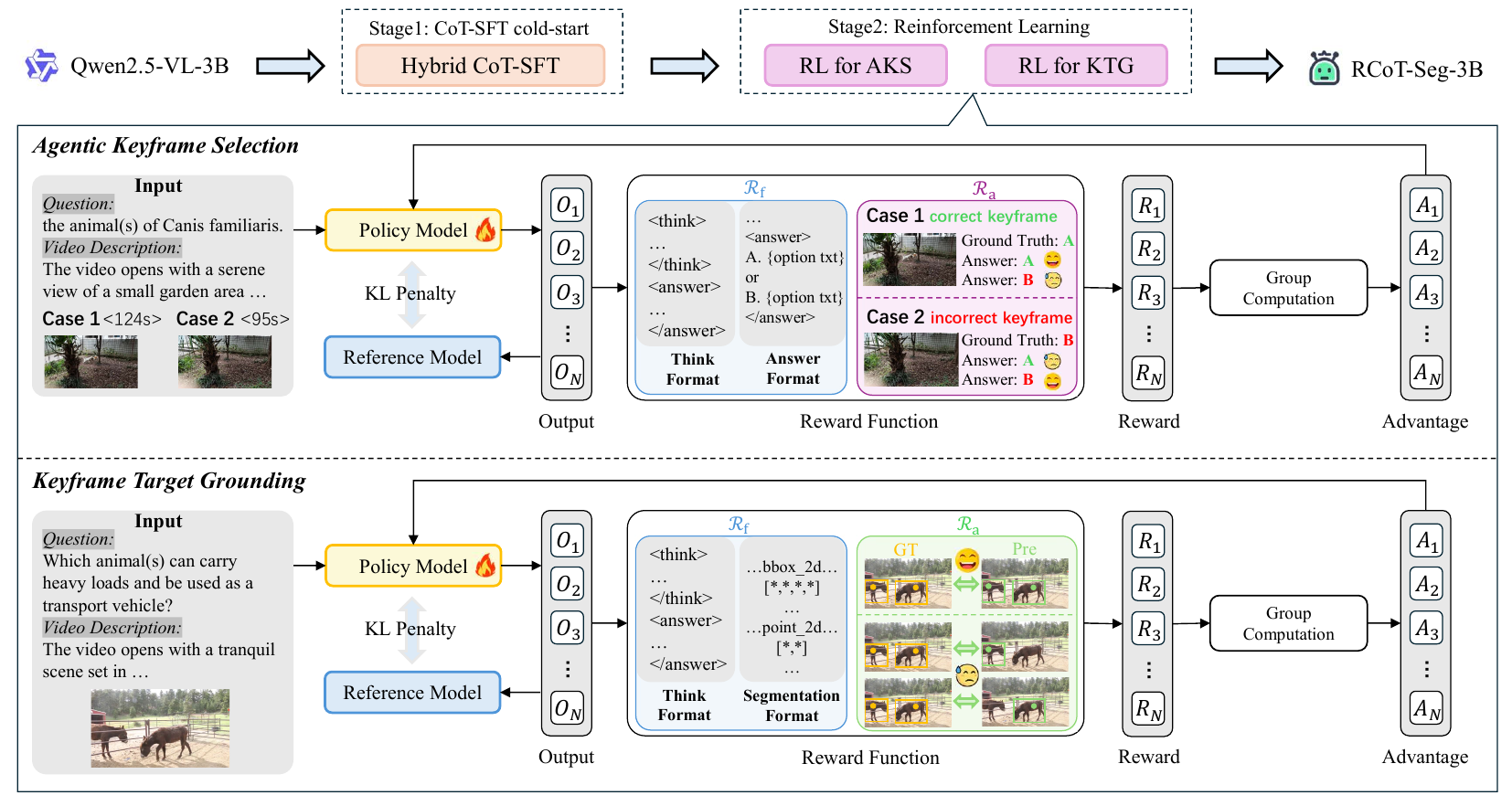}
  }
  \vspace{-2mm}
  \caption{\textbf{Overall training strategy of RCoT-Seg.} Following a CoT-SFT cold-start, we employ GRPO reinforcement learning on the model using reward functions tailored to different tasks.}
  \label{fig:training_strategy}
  \vspace{-5mm}
\end{figure*}

\textbf{Stage 1: CoT-SFT cold-start training.} To stabilize reinforcement learning and avoid cold-start instability, we first construct a compact Chain-of-Thought dataset and perform cold-start supervised fine-tuning (CoT-SFT) on Qwen2.5-VL-3B, which serves as the initial RL actor.

We use Qwen2.5-VL-7B in a controlled data-generation pipeline to obtain step-wise reasoning traces.
First, we define the keyframe as the frame whose annotated mask has the largest area. A structured CoT prompt guides the model to first analyze the scene, then describe the referred object(s), and finally localize them. Each instance is annotated with a bounding box and the center of its maximum inscribed circle. The prompt–annotation pair is then fed to Qwen2.5-VL-7B to produce the reasoning chain.
For keyframe selection, we cast selection as a two-option decision: \emph{Option A} is a valid keyframe instantiated as above, and \emph{Option B} is sampled from frames with zero mask area. A fixed CoT prompt elicits reasoning chains, and we retain only those whose final option matches the ground truth.

Reasoning traces are wrapped in \texttt{<think>} \dots \texttt{</think>} and final answers in \texttt{<answer>} \dots \texttt{</answer>}.
We sample from Ref-YouTube-VOS, MeViS, and ReVOS with a single–multi target ratio of $3{:}1$ and an A–B ratio of $1{:}1$, yielding \textbf{RCoT-Seg-SFT-28k}: $20$k samples for KTG and $8$k for AKS.
To provide a stable initialization for subsequent GRPO training, we then jointly fine-tune Qwen2.5-VL-3B on RCoT-Seg-SFT-28k across both tasks using an autoregressive next-token objective with cross-entropy loss.

\noindent\textbf{Stage 2: GRPO training.} As illustrated in \cref{fig:training_strategy}, for each query $q$, the policy samples $N$ responses $\{o_1,\dots,o_N\}$. Each response receives a verifiable, task-aligned reward $\mathcal{R}$, and we apply group-wise normalization to compute advantages (Eq.~\ref{eq:grpo-adv}). Then, a KL penalty to a reference policy (the CoT-SFT model) stabilizes updates (Eq.~\ref{eq:grpo-obj}). 

We proportionally sample 4,000 and 6,000 instances different from those in SFT to construct the \textbf{RCoT-Seg-GRPO-AKS-4k} and \textbf{RCoT-Seg-GRPO-KTG-6k} datasets for AKS and KTG, respectively.
For either task, the total reward is
\begin{equation}
\mathcal{R}=\lambda_f\,\mathcal{R}_f+\lambda_a\,\mathcal{R}_a,
\end{equation}
and group-relative advantages are computed as
\begin{equation}
A_i=\frac{\mathcal{R}_i-\operatorname{mean}\left(\{\mathcal{R}_1,\dots,\mathcal{R}_N\}\right)}
{\operatorname{std}\left(\{\mathcal{R}_1,\dots,\mathcal{R}_N\}\right)}.
\label{eq:grpo-adv}
\end{equation}
The GRPO objective~\cite{shao2024deepseekmath} is
\begin{equation}
\begin{aligned}
\mathcal{J}_{\mathrm{GRPO}}(\theta) = \mathbb{E}_{q, \{o_i\}\sim\pi_{\theta_{\mathrm{old}}}} \Bigg[ \frac{1}{N} \sum_{i=1}^{N} \Bigg( \min \Bigg( \frac{\pi_{\theta}(o_i \mid q)}{\pi_{\theta_{\mathrm{old}}}(o_i \mid q)} A_i, \\
\mathrm{clip} \left( \frac{\pi_{\theta}(o_i \mid q)}{\pi_{\theta_{\mathrm{old}}}(o_i \mid q)}, 1-\epsilon, 1+\epsilon \right) A_i \Bigg) \Bigg) - \beta \mathbb{D}_{\mathrm{KL}}(\pi_{\theta} \| \pi_{\mathrm{ref}}) \Bigg]
\end{aligned}
\label{eq:grpo-obj}
\end{equation}
where $\pi_{\mathrm{ref}}$ is the CoT-SFT model and $\beta$ controls the KL strength.

\section{Experiments}

\begin{table*}
    \centering
    \caption{\textbf{Performance comparison on VRS datasets.} The best and second-best results are shown in \textcolor{red}{\textbf{red}} and \textcolor{blue}{\textbf{blue}} color. 
    }
    \label{tab:vrs}
    \rowcolors{2}{gray!25}{white}
    \resizebox{\linewidth}{!}{
        \begin{tabular}{l|c|ccc|ccc|ccc|ccc}
    \toprule

    \multirow{2}{*}{\textbf{Method}} & \multirow{2}{*}{\textbf{Venue}} & \multicolumn{3}{c|}{\textbf{ReVOS / referring}} & \multicolumn{3}{c|}{\textbf{ReVOS / reasoning}} & \multicolumn{3}{c|}{\textbf{ReVOS / overall}} & \multicolumn{3}{c}{\textbf{ReasonVOS}} \\
    \cmidrule(lr){3-5} \cmidrule(lr){5-8} \cmidrule(lr){9-11} \cmidrule(lr){12-14}
  & & $\mathcal{J}\&\mathcal{F}$ & $\mathcal{J}$ & $\mathcal{F}$ & $\mathcal{J}\&\mathcal{F}$ & $\mathcal{J}$ & $\mathcal{F}$ & $\mathcal{J}\&\mathcal{F}$ & $\mathcal{J}$ & $\mathcal{F}$ & $\mathcal{J}\&\mathcal{F}$ & $\mathcal{J}$ & $\mathcal{F}$ \\
    \midrule
    \multicolumn{14}{c}{\textit{Previous models using <seg> tokens for mask prediction via SFT training}}\\
    \midrule

    VISA-7B~\cite{yan2024visa} & ECCV2024 & 50.9 &49.2 &52.6 & 43.0 &40.6 &45.4 & 46.9 &44.9 &49.0 &  -& -&-\\
    VISA-13B~\cite{yan2024visa} & ECCV2024 & 57.4 &55.6 &59.1 & 44.3 &42.0 &46.7 & 50.9 &48.8 &52.9&  -&- &-\\
    VideoLISA-3.8B~\cite{videolisa} & NeurIPS2024 & - &- &- &  -& -&- &  -&- &- &  47.5 &45.1 &49.9\\
    InstructSeg-3B~\cite{wei2025instructseg} & ICCV2025 & 57.0 &54.8 &59.2 & 51.9 &49.2 &54.7 & 54.5 &52.0 &56.9& - & -&-\\
    HyperSeg-3B~\cite{hyperseg} & CVPR2025 & 58.5 &56.0 &60.9 & 53.0 &50.2 &55.8 & 55.7 &53.1 &58.4&  -&- &-\\
    GLUS-7B~\cite{lin2025glus} & CVPR2025 & 58.3 &56.0 &60.7 & 51.4 &48.8 &53.9 & 54.8 &52.4 &57.3& 49.9 &47.5 &52.4\\
    VRS-HQ-7B~\cite{VRS-HQ} & CVPR2025 & 62.1 &59.8 &64.5 & 56.1 &53.5 &58.7 & 59.1 &56.6 &61.6 & -& -&-\\
    VRS-HQ-13B~\cite{VRS-HQ} & CVPR2025 & \textcolor{blue}{\textbf{63.3}} &\textcolor{blue}{\textbf{61.1}} &65.5 & \textcolor{blue}{\textbf{56.8}} &\textcolor{blue}{\textbf{54.1}} &59.4 & \textcolor{blue}{\textbf{60.0}} &\textcolor{blue}{\textbf{57.6}} &62.5 &  -& -&-\\

    \midrule
    \multicolumn{14}{c}{\textit{Latest models via GRPO training}}\\
    \midrule
    Omni-R1-8B~\cite{zhong2025omni} & NeurIPS2025 &  61.6 &- &- & 50.7 &- &- & 56.2 &- &- & - &- &- \\
    Veason-R1-3B~\cite{gong2025reinforcing} & CVPR2026 & 63.0 &60.3 &\textcolor{blue}{\textbf{65.6}} & \textcolor{blue}{\textbf{56.8}} &53.6 &\textcolor{blue}{\textbf{60.0}} & 59.9 &56.9 &\textcolor{blue}{\textbf{62.8}}& \textcolor{blue}{\textbf{55.2}} &\textcolor{blue}{\textbf{51.8}} &\textcolor{blue}{\textbf{58.5}}\\

    \textbf{RCoT-Seg-3B} & -
    & \textcolor{red}{\textbf{64.3}}~\textbf{(+1.0)}
    & \textcolor{red}{\textbf{61.6}}
    & \textcolor{red}{\textbf{66.9}}
    & \textcolor{red}{\textbf{58.2}}~\textbf{(+1.4)}
    & \textcolor{red}{\textbf{54.9}}
    & \textcolor{red}{\textbf{61.6}}
    & \textcolor{red}{\textbf{61.2}}~\textbf{(+1.2)}
    & \textcolor{red}{\textbf{58.2}}
    & \textcolor{red}{\textbf{64.3}}
    & \textcolor{red}{\textbf{58.2}}~\textbf{(+3.0)}
    & \textcolor{red}{\textbf{54.6}}
    & \textcolor{red}{\textbf{61.8}}\\
    \bottomrule
  \end{tabular}}
  \vspace{-2mm}
\end{table*}

\subsection{Datasets and metrics}
\textbf{Training datasets.} We select training samples from the Ref-YouTube-VOS~\cite{seo2020urvos}, MeViS~\cite{ding2023mevis}, and ReVOS~\cite{yan2024visa} training datasets according to a predefined ratio of single-multi and A-B. During the SFT stage, we adapt the Qwen2.5-VL-3B~\cite{bai2025qwen2} model using the hybrid chain-of-thought dataset RCoT-Seg-SFT-28k, which is specifically designed for AKS and KTG tasks. In the GRPO training phase, we construct the RCoT-Seg-GRPO-AKS-4k and RCoT-Seg-GRPO-KTG-6k datasets for AKS and KTG, respectively.

\noindent\textbf{Evaluation datasets.} To comprehensively evaluate the performance of RCoT-Seg, we perform testing of RCoT-Seg across several benchmarks. These include two VRS datasets: ReVOS and ReasonVOS~\cite{videolisa}, as well as three RVOS datasets: DAVIS17~\cite{pont20172017}, Ref-YouTube-VOS, and MeViS~\cite{ding2023mevis}.

\noindent\textbf{Evaluation metrics.} Following the evaluation in prior work~\cite{yan2024visa}, we employ three metrics: region similarity ($\mathcal{J}$), contour accuracy ($\mathcal{F}$), and their composite measure ($\mathcal{J}\&\mathcal{F}$). The $\mathcal{J}$ metric computes the intersection-over-union (IoU) between the predicted mask sequence and the ground truth, while the $\mathcal{F}$ metric assesses boundary alignment precision based on contour matching.
\subsection{Implementation details}
During the supervised fine-tuning (SFT) stage, we utilize the LLaMA-Factory framework~\cite{zheng2024llamafactory} to fine-tune the Qwen2.5-VL-3B model with LoRA~\cite {hu2022lora}(rank=8) while freezing all other parameters. The training configuration employs a learning rate of $1 \times 10^{-4}$, cosine annealing scheduling, and 8-step gradient accumulation, conducting one training epoch on our chain-of-thought dataset RCoT-Seg-SFT-28k. In the reinforcement learning training phase, we implement the VERL framework~\cite{sheng2025hybridflow} with a global batch size of 16, sampling 8 responses per input prompt to facilitate preference optimization. The model is trained for 500 and 1000 steps for AKS and KTG, respectively, using a learning rate of $1 \times 10^{-6}$. All experiments are conducted on 4 NVIDIA L20 GPUs.

\subsection{Comparison results}
\textbf{Video reasoning segmentation.} \cref{tab:vrs} presents a detailed performance comparison between RCoT-Seg and previous models~\cite{videolisa,yan2024visa,lin2025glus,VRS-HQ} trained via SFT using <seg> tokens for mask prediction, as well as the latest GRPO-trained model~\cite{zhong2025omni,gong2025reinforcing}. Our method exhibits competitive performance compared to the
leading state-of-the-art methods. Specifically, despite being fine-tuned on a limited number of samples, RCoT-Seg surpasses the previous SOTA model VRS-HQ-13B~\cite{VRS-HQ} by \textbf{1.2}\% in $\mathcal{J}\&\mathcal{F}$ on ReVOS, and outperforms GLUS-7B~\cite{lin2025glus} by \textbf{8.3}\% in $\mathcal{J}\&\mathcal{F}$ on ReasonVOS, which stems primarily from the improved reasoning ability of our method.  
Furthermore, RCoT-Seg exceeds Omni-R1~\cite{zhong2025omni} by approximately \textbf{5.0}\% on ReVOS, demonstrating the incorporation of multi-task reasoning to guide precise segmentation.
Moreover, RCoT-Seg achieves higher performance than the GRPO-based method, Veason-R1~\cite{gong2025reinforcing}, on both ReVOS (\textbf{+1.3}\%) and ReasonVOS datasets (\textbf{+3.0}\%), which underscores the efficacy of the agentic keyframe selection mechanism.

\begin{table*}
    \centering
    \caption{\textbf{Performance comparison on RVOS datasets.} The best and second-best results are shown in \textcolor{red}{\textbf{red}} and \textcolor{blue}{\textbf{blue}} color.}
    \footnotesize
    \label{tab:rvos}
    \rowcolors{2}{gray!25}{white}
    \resizebox{0.95\linewidth}{!}{
        \begin{tabular}{l|c|ccc|ccc|ccc}
    \toprule
    
    \rowcolors{2}{gray!25}{white}
    \multirow{2}{*}{\textbf{Method}} & \multirow{2}{*}{\textbf{MLLM}} & \multicolumn{3}{c|}{\textbf{DAVIS17}} & \multicolumn{3}{c|}{\textbf{Ref-YouTube-VOS}} & \multicolumn{3}{c}{\textbf{MeViS}}\\
    \cmidrule(lr){3-5} \cmidrule(lr){6-8} \cmidrule(lr){9-11}
 & & $\mathcal{J}\&\mathcal{F}$ & $\mathcal{J}$ & $\mathcal{F}$ & $\mathcal{J}\&\mathcal{F}$ & $\mathcal{J}$ & $\mathcal{F}$ & $\mathcal{J}\&\mathcal{F}$ & $\mathcal{J}$ & $\mathcal{F}$ \\
    \midrule

    Track-GPT-7B~\cite{zhu2023tracking} &LLaVA-7B & 63.2 &59.4 &67.0 & 56.4 &55.3 &57.4 & 40.1 &37.6 &42.6\\
    Track-GPT-13B~\cite{zhu2023tracking} &LLaVA-13B & 66.5 &62.7 &70.4 & 59.5 &58.1 &60.8 & 41.2 &39.2 &43.1\\
    VISA-7B~\cite{yan2024visa} & Chat-UniVi-7B & 69.4 &66.3 &72.5 & 61.5 &59.8 &63.2 & 43.5 &40.7 &46.3\\
    VISA-13B~\cite{yan2024visa} & Chat-UniVi-13B & 70.4 &67.0 &73.8 & 63.0 &61.4 &64.7 & 44.5 &41.8 &47.1\\
    VideoLISA-3.8B~\cite{videolisa}& LLaVA-Phi-3-V & 68.8 &59.4 &64.9 & 63.7 &61.7 &65.7 & 44.4 &41.3 &47.6\\
    GLUS-7B~\cite{lin2025glus}& LLaVA-7B & - &- &- & 67.3 &65.5 &69.0 & \textcolor{blue}{\textbf{51.3}} &\textcolor{blue}{\textbf{48.5}} &\textcolor{blue}{\textbf{54.2}}\\
    VRS-HQ-13B~\cite{VRS-HQ} & Chat-UniVi-13B & \textcolor{blue}{\textbf{74.4}} &\textcolor{blue}{\textbf{71.0}} &\textcolor{blue}{\textbf{77.9}} & \textcolor{blue}{\textbf{71.0}} &\textcolor{blue}{\textbf{69.0}} &\textcolor{blue}{\textbf{73.1}} & 50.9 &48.0 &53.7\\

    \midrule
    Veason-R1-3B~\cite{gong2025reinforcing} & Qwen2.5-VL-3B & - &- &- & - &- &- & 51.2 &48.2 &\textcolor{blue}{\textbf{54.2}}\\

    \textbf{RCoT-Seg(ours)}& Qwen2.5-VL-3B 
& \textcolor{red}{\textbf{76.1}}~\textbf{(+1.7)}
& \textcolor{red}{\textbf{72.2}}
& \textcolor{red}{\textbf{80.0}}
& \textcolor{red}{\textbf{72.5}}~\textbf{(+1.5)}
& \textcolor{red}{\textbf{70.5}}
& \textcolor{red}{\textbf{74.6}}
& \textcolor{red}{\textbf{53.7}}~\textbf{(+2.4)}
& \textcolor{red}{\textbf{50.7}}
& \textcolor{red}{\textbf{56.7}}\\
    
    \bottomrule
  \end{tabular}}
\end{table*}

\begin{table*}[htbp]
\centering

\begin{minipage}[t]{0.48\textwidth}
\centering
\caption{Ablation on the proposed components. VDG, KFG and AKS denote Video Description Generation, Keyframe Generation, and Agentic Keyframe Selection, respectively. When KFG is not used, we select the first frame as the keyframe.}
\label{tab:ablation_component}
\resizebox{\linewidth}{!}{
\begin{tabular}{c|c|c|c|c|c|c}
\toprule
VDG & KFG & AKS & Ref-YT-VOS & MeViS & ReVOS & ReasonVOS \\
\midrule
    &   &   & 68.9 & 48.9 & 57.1 & 53.9 \\
\checkmark &   &   & 69.1 & 49.8 & 57.7 & 54.9 \\
    & \checkmark &   & 70.5 & 51.7 & 59.5 & 54.4 \\
\checkmark & \checkmark &   & 70.8 & 52.2 & 60.3 & 55.7 \\
\checkmark & \checkmark & \checkmark & \textbf{72.5} & \textbf{53.7} & \textbf{61.2} & \textbf{58.2} \\
\bottomrule
\end{tabular}}
\end{minipage}
\hfill
\begin{minipage}[t]{0.48\textwidth}
\centering
\caption{Ablation on the maximum iteration count $\lambda$ for agentic keyframe selection. $\lambda=0$ indicates that AKS is not used. $\lambda=5$ achieves the best performance, we select $\lambda=5$ in default.}
\label{tab:ablation_agent}
\resizebox{\linewidth}{!}{
\begin{tabular}{c|c|c|c|c|c}
\toprule
$\lambda$ & DAVIS17 & Ref-YT-VOS & MeViS & ReVOS & ReasonVOS \\
\midrule
0 & 74.1 & 70.8 & 52.2 & 60.3 & 55.7  \\
2 & 74.6 & 71.6 & 53.2 & 60.8 & 57.2 \\
3 & 75.3 & 72.2 & 53.6 & 60.9 & 57.8 \\
\textbf{5} & \textbf{76.1} & \textbf{72.5} & 53.7 & \textbf{61.2} & \textbf{58.2}  \\
7 & 76.0 & 72.5 & \textbf{53.8} & 61.2 & 58.1 \\
10 & 76.1 & 72.4 & 53.7 & 61.2 & 58.2 \\
\bottomrule
\end{tabular}}
\end{minipage}

\vspace{3mm}

\begin{minipage}[t]{0.48\textwidth}
\centering
\caption{Ablation on different training strategies. $SFT^{A}$ denotes performing SFT on the agentic keyframe selection task, $SFT^{G}$ denotes performing SFT on the keyframe target grounding task, and $SFT^{A,G}$ denotes performing SFT on both tasks. }
\label{tab:ablation_training}
\resizebox{\linewidth}{!}{
\begin{tabular}{c|ccc|ccc}
\toprule
\multirow{2}{*}{Training Strategy} & \multicolumn{3}{c|}{Referring} & \multicolumn{3}{c}{Reasoning} \\
\cmidrule(lr){2-4} \cmidrule(lr){5-7} 
& $\mathcal{J}\&\mathcal{F}$ & $\mathcal{J}$ & $\mathcal{F}$ & $\mathcal{J}\&\mathcal{F}$ & $\mathcal{J}$ & $\mathcal{F}$ \\
\midrule
Qwen2.5-VL-3B & 14.7 & 11.3 & 18.2 & 13.7 & 10.5 & 17.0 \\
\midrule
$SFT^{A,G}$ & 59.1 & 56.4 & 61.8 & 51.1 & 48.0 & 54.1  \\
$GRPO^{A,G}$ & 62.8 & 60.1 & 65.5 & 55.5 & 52.3 & 58.8  \\
$SFT^{A,G}+GRPO^{G}$ & 63.8 & 60.8 & 66.8 & 56.3 & 53.1 & 59.5  \\
$SFT^{G}+GRPO^{A,G}$ & 63.0 & 60.3 & 65.6 & 57.0 & 54.0 & 60.0  \\
$SFT^{A,G}+GRPO^{A,G}$ &\textbf{64.3} &\textbf{61.6} &\textbf{66.9} &\textbf{58.2} &\textbf{54.9} &\textbf{61.6}  \\
\bottomrule
\end{tabular}}
\end{minipage}
\hfill
\begin{minipage}[t]{0.48\textwidth}
\centering
\caption{Ablation on architecture. The separated architecture means we adopt CoT-SFT cold-start GRPO to train two separated models for completing the AKS and VTG tasks, while utilizing Qwen2.5-VL-3B to handle the VDG and KFG tasks. The unified architecture refers to the proposed RCoT-Seg.}
\label{tab:ablation_uni}
\resizebox{\linewidth}{!}{
\begin{tabular}{c|ccc|ccc}
\toprule
\multirow{2}{*}{Architecture} & \multicolumn{3}{c|}{Referring} & \multicolumn{3}{c}{Reasoning} \\
\cmidrule(lr){2-4} \cmidrule(lr){5-7} 
& $\mathcal{J}\&\mathcal{F}$ & $\mathcal{J}$ & $\mathcal{F}$ & $\mathcal{J}\&\mathcal{F}$ & $\mathcal{J}$ & $\mathcal{F}$ \\
\midrule
Separated & 62.5 & 59.6 & 65.4 & 55.6 & 52.1 & 59.1 \\
Unified  &\textbf{64.3} &\textbf{61.6} &\textbf{66.9} &\textbf{58.2} &\textbf{54.9} &\textbf{61.6}  \\
\bottomrule
\end{tabular}}
\end{minipage}

\end{table*}

\noindent\textbf{Referring VOS.} \cref{tab:rvos} presents a performance comparison between RCoT-Seg and state-of-the-art RVOS methods across three benchmark datasets.
The proposed RCoT-Seg surpasses VRS-HQ-13B~\cite{VRS-HQ} by \textbf{1.7}\% in $\mathcal{J}\&\mathcal{F}$ on DAVIS17 and  by \textbf{1.5}\% in $\mathcal{J}\&\mathcal{F}$ on Ref-YouTube-VOS, while utilizing fewer parameters. 
On the more complex MeViS dataset, RCoT-Seg outperforms GLUS-7B~\cite{lin2025glus} by \textbf{2.4}\% in $\mathcal{J}\&\mathcal{F}$, highlighting the effectiveness in handling complex scenarios. 
Furthermore, RCoT-Seg exceeds Veason-R1~\cite{gong2025reinforcing} by \textbf{2.5}\% on the MeViS dataset, indicating the efficacy of the agentic keyframe selection mechanism.

\subsection{Ablation Studies}
\textbf{Effect of the proposed components.} As shown in \cref{tab:ablation_component}, the adoption of our Video Description Generation, Keyframe Generation, and Agentic Keyframe Selection modules leads to substantial improvements (\textbf{+3.6} on Ref-YouTube-VOS, \textbf{+4.8} on MeViS, \textbf{+4.1} on ReVOS, \textbf{+4.3} on ReasonVOS).

\noindent\textbf{Ablation on the maximum iteration count for AKS.} \cref{tab:ablation_agent} shows the impact of the maximum iteration count $\lambda$ for AKS on the performance. The results indicate that a count of 5 achieves a favorable balance, yielding strong performance while reducing redundancy. Consequently, a threshold of 5 is adopted for our model during inference. We also analyze the inference time of this component. Please see Appendix \ref{appendix:time-analysis} for details.

\noindent\textbf{Effect of our training strategy.} The results in \cref{tab:ablation_training} validates the efficacy of our training strategy. When SFT and GRPO are applied individually to the two tasks, the performance on ReVOS decreases by \textbf{5.2}\% and \textbf{1.5}\% in $\mathcal{J}\&\mathcal{F}$ on the referring subset, and by \textbf{7.1}\% and \textbf{2.7}\% in $\mathcal{J}\&\mathcal{F}$ on the reasoning subset, respectively. Furthermore, when GRPO and SFT are omitted individually for the AKS task, the performance on ReVOS decreases by \textbf{0.5}\% and \textbf{1.3}\% in $\mathcal{J}\&\mathcal{F}$ on the referring subset, and by \textbf{1.9}\% and \textbf{1.2}\% in $\mathcal{J}\&\mathcal{F}$ on the reasoning subset, respectively. This demonstrates the effectiveness of our multi-task, multi-stage training strategy.

\begin{figure*}
  \centering
  \resizebox{\linewidth}{!}{
    \includegraphics{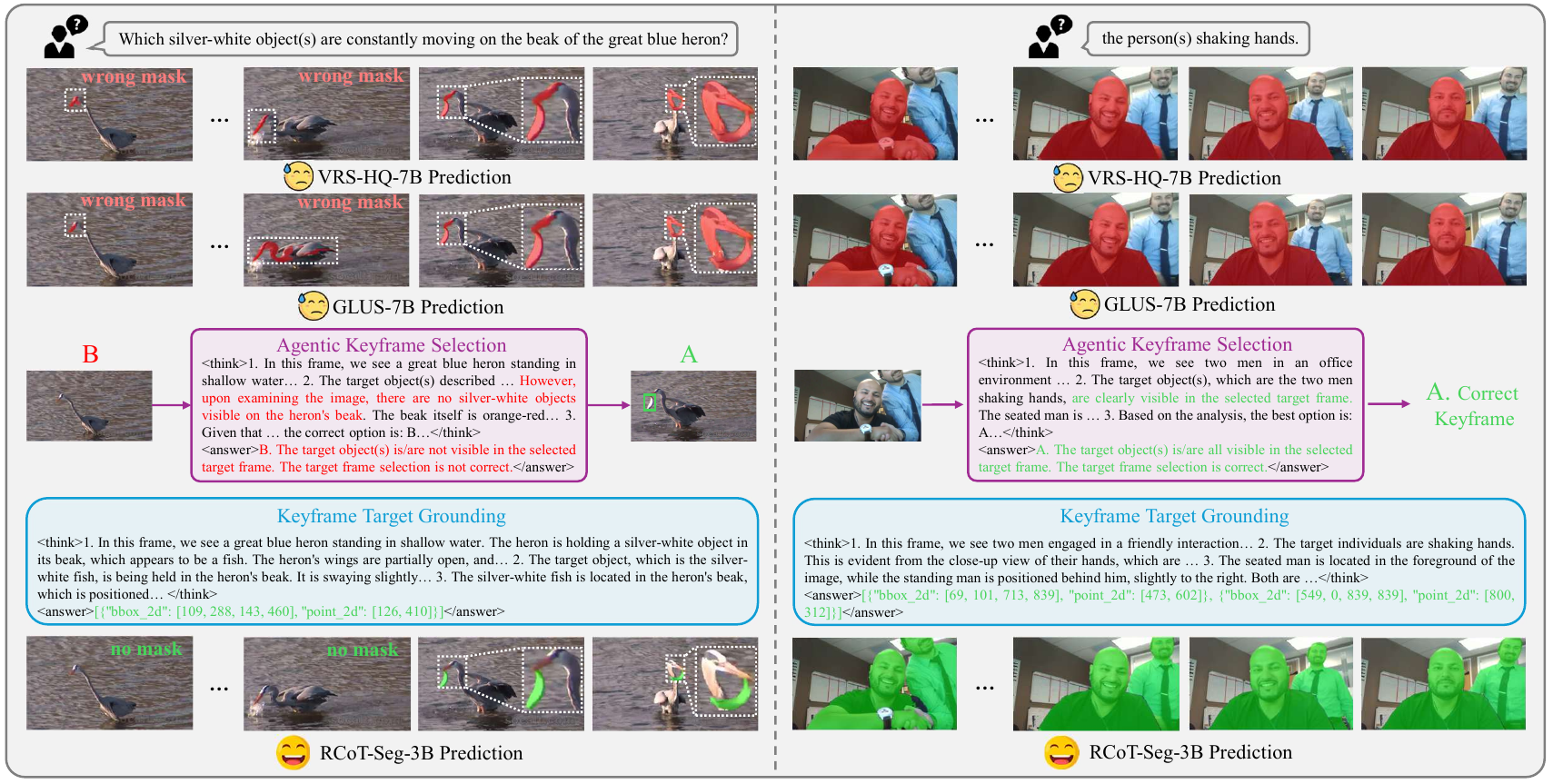}
  }
  \caption{\textbf{Qualitative comparison between RCoT-Seg and current SOTA methods.} Our method successfully identify and accurately segment targets in complex scenarios involving small-scale objects and multi-object interactions.}
  \label{fig:visual-results}
\end{figure*}

\noindent\textbf{Effect of our unified model.} We attempt to address VRS under the separated architecture, which allocates different tasks to separated models. The results in \cref{tab:ablation_uni} indicate that this separated architecture underperforms the proposed unified RCoT-Seg model by \textbf{1.8}\% in $\mathcal{J}\&\mathcal{F}$ on the referring subset and by \textbf{2.6}\% in $\mathcal{J}\&\mathcal{F}$ on the reasoning subset, demonstrating the efficacy of our unified architecture.


\subsection{Qualitative comparison}
\cref{fig:visual-results} qualitatively compares RCoT-Seg with VRS-HQ-7B~\cite{VRS-HQ} and GLUS-7B~\cite{lin2025glus} in complex scenarios. For the left example, where the target object is small and appears only in a subset of frames, RCoT-Seg employs AKS to reselect the correct keyframe, thereby achieving a better segmentation result. In the right case, it is a multi-object scenario (two people shaking hands). Both VRS-HQ-7B~\cite{VRS-HQ} and GLUS-7B~\cite{lin2025glus} segment only one of them, whereas our RCoT-Seg successfully segments all target objects.

\section{Conclusion}
In this work, we present RCoT-Seg, a unified model that accomplishes VRS through a sequential pipeline of temporal video reasoning and keyframe target perception trained with reinforcement learning. 
Unlike sampling-based paradigms, we propose an agentic keyframe selection to improve keyframe quality with a simple self-evolution mechanism.
The KTP stage operates on the selected keyframe guided by a compact video description, which transfers long-horizon context to pixel-level decoding. 
To handle multi-object scenarios, we design a Hungarian-based matching reward that stabilizes training and strengthens object-wise consistency.
The model is cold-started with the proposed hybrid CoT supervision and then optimized across tasks using GRPO, aligning reasoning traces with downstream segmentation behavior.
Extensive experiments on standard benchmarks show that RCoT-Seg achieves state-of-the-art performance in video reasoning segmentation and video referring segmentation, validating the effectiveness of the global-to-local design, the agentic selection mechanism, and the explicit Chain-of-Thought guidance in VRS.

\bibliography{references}

@article{hyperseg,
  title={HyperSeg: Towards Universal Visual Segmentation with Large Language Model},
  author={Wei, Cong and Zhong, Yujie and Tan, Haoxian and Liu, Yong and Zhao, Zheng and Hu, Jie and Yang, Yujiu},
  journal={arXiv preprint arXiv:2411.17606},
  year={2024}
}

@inproceedings{seo2020urvos,
  title={Urvos: Unified referring video object segmentation network with a large-scale benchmark},
  author={Seo, Seonguk and Lee, Joon-Young and Han, Bohyung},
  booktitle={Computer Vision--ECCV 2020: 16th European Conference, Glasgow, UK, August 23--28, 2020, Proceedings, Part XV 16},
  pages={208--223},
  year={2020},
  organization={Springer}
}

@inproceedings{ding2023mevis,
  title={Mevis: A large-scale benchmark for video segmentation with motion expressions},
  author={Ding, Henghui and Liu, Chang and He, Shuting and Jiang, Xudong and Loy, Chen Change},
  booktitle={Proceedings of the IEEE/CVF international conference on computer vision},
  pages={2694--2703},
  year={2023}
}

@inproceedings{yan2024visa,
  title={Visa: Reasoning video object segmentation via large language models},
  author={Yan, Cilin and Wang, Haochen and Yan, Shilin and Jiang, Xiaolong and Hu, Yao and Kang, Guoliang and Xie, Weidi and Gavves, Efstratios},
  booktitle={European Conference on Computer Vision},
  pages={98--115},
  year={2024},
  organization={Springer}
}

@inproceedings{VRS-HQ,
  title={The devil is in temporal token: High quality video reasoning segmentation},
  author={Gong, Sitong and Zhuge, Yunzhi and Zhang, Lu and Yang, Zongxin and Zhang, Pingping and Lu, Huchuan},
  booktitle={Proceedings of the Computer Vision and Pattern Recognition Conference},
  pages={29183--29192},
  year={2025}
}

@article{zheng2024villa,
  title={ViLLa: Video Reasoning Segmentation with Large Language Model},
  author={Zheng, Rongkun and Qi, Lu and Chen, Xi and Wang, Yi and Wang, Kun and Qiao, Yu and Zhao, Hengshuang},
  journal={arXiv preprint arXiv:2407.14500},
  year={2024}
}

@article{videolisa,
  title={One token to seg them all: Language instructed reasoning segmentation in videos},
  author={Bai, Zechen and He, Tong and Mei, Haiyang and Wang, Pichao and Gao, Ziteng and Chen, Joya and Liu, Lei and Zhang, Zheng and Shou, Mike Z},
  journal={Advances in Neural Information Processing Systems},
  volume={37},
  pages={6833--6859},
  year={2024}
}

@article{gong2025reinforcing,
  title={Reinforcing video reasoning segmentation to think before it segments},
  author={Gong, Sitong and Zhang, Lu and Zhuge, Yunzhi and Jia, Xu and Zhang, Pingping and Lu, Huchuan},
  journal={arXiv preprint arXiv:2508.11538},
  year={2025}
}

@inproceedings{lin2025glus,
  title={Glus: Global-local reasoning unified into a single large language model for video segmentation},
  author={Lin, Lang and Yu, Xueyang and Pang, Ziqi and Wang, Yu-Xiong},
  booktitle={Proceedings of the Computer Vision and Pattern Recognition Conference},
  pages={8658--8667},
  year={2025}
}

@book{sutton1998reinforcement,
  title={Reinforcement learning: An introduction},
  author={Sutton, Richard S and Barto, Andrew G and others},
  volume={1},
  number={1},
  year={1998},
  publisher={MIT press Cambridge}
}

@article{jaech2024openai,
  title={Openai o1 system card},
  author={Jaech, Aaron and Kalai, Adam and Lerer, Adam and Richardson, Adam and El-Kishky, Ahmed and Low, Aiden and Helyar, Alec and Madry, Aleksander and Beutel, Alex and Carney, Alex and others},
  journal={arXiv preprint arXiv:2412.16720},
  year={2024}
}

@article{guo2025deepseek,
  title={Deepseek-r1: Incentivizing reasoning capability in llms via reinforcement learning},
  author={Guo, Daya and Yang, Dejian and Zhang, Haowei and Song, Junxiao and Zhang, Ruoyu and Xu, Runxin and Zhu, Qihao and Ma, Shirong and Wang, Peiyi and Bi, Xiao and others},
  journal={arXiv preprint arXiv:2501.12948},
  year={2025}
}

@article{liu2025seg,
  title={Seg-zero: Reasoning-chain guided segmentation via cognitive reinforcement},
  author={Liu, Yuqi and Peng, Bohao and Zhong, Zhisheng and Yue, Zihao and Lu, Fanbin and Yu, Bei and Jia, Jiaya},
  journal={arXiv preprint arXiv:2503.06520},
  year={2025}
}

@article{liu2025visionreasoner,
  title={VisionReasoner: Unified Visual Perception and Reasoning via Reinforcement Learning},
  author={Liu, Yuqi and Qu, Tianyuan and Zhong, Zhisheng and Peng, Bohao and Liu, Shu and Yu, Bei and Jia, Jiaya},
  journal={arXiv preprint arXiv:2505.12081},
  year={2025}
}

@article{liu2025visual,
  title={Visual-rft: Visual reinforcement fine-tuning},
  author={Liu, Ziyu and Sun, Zeyi and Zang, Yuhang and Dong, Xiaoyi and Cao, Yuhang and Duan, Haodong and Lin, Dahua and Wang, Jiaqi},
  journal={arXiv preprint arXiv:2503.01785},
  year={2025}
}

@article{wang2025pixelthink,
  title={PixelThink: Towards Efficient Chain-of-Pixel Reasoning},
  author={Wang, Song and Fang, Gongfan and Kong, Lingdong and Li, Xiangtai and Xu, Jianyun and Yang, Sheng and Li, Qiang and Zhu, Jianke and Wang, Xinchao},
  journal={arXiv preprint arXiv:2505.23727},
  year={2025}
}

@article{feng2025video,
  title={Video-r1: Reinforcing video reasoning in mllms},
  author={Feng, Kaituo and Gong, Kaixiong and Li, Bohao and Guo, Zonghao and Wang, Yibing and Peng, Tianshuo and Wu, Junfei and Zhang, Xiaoying and Wang, Benyou and Yue, Xiangyu},
  journal={arXiv preprint arXiv:2503.21776},
  year={2025}
}

@article{li2025videochat,
  title={Videochat-r1: Enhancing spatio-temporal perception via reinforcement fine-tuning},
  author={Li, Xinhao and Yan, Ziang and Meng, Desen and Dong, Lu and Zeng, Xiangyu and He, Yinan and Wang, Yali and Qiao, Yu and Wang, Yi and Wang, Limin},
  journal={arXiv preprint arXiv:2504.06958},
  year={2025}
}

@article{wang2025time,
  title={Time-R1: Post-Training Large Vision Language Model for Temporal Video Grounding},
  author={Wang, Ye and Wang, Ziheng and Xu, Boshen and Du, Yang and Lin, Kejun and Xiao, Zihan and Yue, Zihao and Ju, Jianzhong and Zhang, Liang and Yang, Dingyi and others},
  journal={arXiv preprint arXiv:2503.13377},
  year={2025}
}

@article{bai2025univg,
  title={Univg-r1: Reasoning guided universal visual grounding with reinforcement learning},
  author={Bai, Sule and Li, Mingxing and Liu, Yong and Tang, Jing and Zhang, Haoji and Sun, Lei and Chu, Xiangxiang and Tang, Yansong},
  journal={arXiv preprint arXiv:2505.14231},
  year={2025}
}

@article{zhang2025improving,
  title={Improving the reasoning of multi-image grounding in mllms via reinforcement learning},
  author={Zhang, Bob and Li, Haoran and Zhang, Tao and Yan, Cilin and Cai, Jiayin and Hao, Yanbin},
  journal={arXiv preprint arXiv:2507.00748},
  year={2025}
}

@article{fang2025got,
  title={Got: Unleashing reasoning capability of multimodal large language model for visual generation and editing},
  author={Fang, Rongyao and Duan, Chengqi and Wang, Kun and Huang, Linjiang and Li, Hao and Yan, Shilin and Tian, Hao and Zeng, Xingyu and Zhao, Rui and Dai, Jifeng and others},
  journal={arXiv preprint arXiv:2503.10639},
  year={2025}
}

@article{xiao2025mindomni,
  title={Mindomni: Unleashing reasoning generation in vision language models with rgpo},
  author={Xiao, Yicheng and Song, Lin and Chen, Yukang and Luo, Yingmin and Chen, Yuxin and Gan, Yukang and Huang, Wei and Li, Xiu and Qi, Xiaojuan and Shan, Ying},
  journal={arXiv preprint arXiv:2505.13031},
  year={2025}
}

@article{xue2025dancegrpo,
  title={DanceGRPO: Unleashing GRPO on Visual Generation},
  author={Xue, Zeyue and Wu, Jie and Gao, Yu and Kong, Fangyuan and Zhu, Lingting and Chen, Mengzhao and Liu, Zhiheng and Liu, Wei and Guo, Qiushan and Huang, Weilin and others},
  journal={arXiv preprint arXiv:2505.07818},
  year={2025}
}

@article{zhong2025omni,
  title={Omni-R1: Reinforcement Learning for Omnimodal Reasoning via Two-System Collaboration},
  author={Zhong, Hao and Zhu, Muzhi and Du, Zongze and Huang, Zheng and Zhao, Canyu and Liu, Mingyu and Wang, Wen and Chen, Hao and Shen, Chunhua},
  journal={arXiv preprint arXiv:2505.20256},
  year={2025}
}

@article{bai2025qwen2,
  title={Qwen2. 5-vl technical report},
  author={Bai, Shuai and Chen, Keqin and Liu, Xuejing and Wang, Jialin and Ge, Wenbin and Song, Sibo and Dang, Kai and Wang, Peng and Wang, Shijie and Tang, Jun and others},
  journal={arXiv preprint arXiv:2502.13923},
  year={2025}
}

@article{shao2024deepseekmath,
  title={Deepseekmath: Pushing the limits of mathematical reasoning in open language models},
  author={Shao, Zhihong and Wang, Peiyi and Zhu, Qihao and Xu, Runxin and Song, Junxiao and Bi, Xiao and Zhang, Haowei and Zhang, Mingchuan and Li, YK and Wu, Yang and others},
  journal={arXiv preprint arXiv:2402.03300},
  year={2024}
}

@article{ravi2024sam,
  title={Sam 2: Segment anything in images and videos},
  author={Ravi, Nikhila and Gabeur, Valentin and Hu, Yuan-Ting and Hu, Ronghang and Ryali, Chaitanya and Ma, Tengyu and Khedr, Haitham and R{\"a}dle, Roman and Rolland, Chloe and Gustafson, Laura and others},
  journal={arXiv preprint arXiv:2408.00714},
  year={2024}
}

@article{hu2022lora,
  title={Lora: Low-rank adaptation of large language models.},
  author={Hu, Edward J and Shen, Yelong and Wallis, Phillip and Allen-Zhu, Zeyuan and Li, Yuanzhi and Wang, Shean and Wang, Lu and Chen, Weizhu and others},
  journal={ICLR},
  volume={1},
  number={2},
  pages={3},
  year={2022}
}

@article{pont20172017,
  title={The 2017 davis challenge on video object segmentation},
  author={Pont-Tuset, Jordi and Perazzi, Federico and Caelles, Sergi and Arbel{\'a}ez, Pablo and Sorkine-Hornung, Alex and Van Gool, Luc},
  journal={arXiv preprint arXiv:1704.00675},
  year={2017}
}

@article{zheng2024llamafactory,
  title={Llamafactory: Unified efficient fine-tuning of 100+ language models},
  author={Zheng, Yaowei and Zhang, Richong and Zhang, Junhao and Ye, Yanhan and Luo, Zheyan and Feng, Zhangchi and Ma, Yongqiang},
  journal={arXiv preprint arXiv:2403.13372},
  year={2024}
}

@inproceedings{sheng2025hybridflow,
  title={Hybridflow: A flexible and efficient rlhf framework},
  author={Sheng, Guangming and Zhang, Chi and Ye, Zilingfeng and Wu, Xibin and Zhang, Wang and Zhang, Ru and Peng, Yanghua and Lin, Haibin and Wu, Chuan},
  booktitle={Proceedings of the Twentieth European Conference on Computer Systems},
  pages={1279--1297},
  year={2025}
}

@inproceedings{xia2024gsva,
  title={Gsva: Generalized segmentation via multimodal large language models},
  author={Xia, Zhuofan and Han, Dongchen and Han, Yizeng and Pan, Xuran and Song, Shiji and Huang, Gao},
  booktitle={Proceedings of the IEEE/CVF Conference on Computer Vision and Pattern Recognition},
  pages={3858--3869},
  year={2024}
}

@inproceedings{wei2025instructseg,
  title={Instructseg: Unifying instructed visual segmentation with multi-modal large language models},
  author={Wei, Cong and Zhong, Yujie and Tan, Haoxian and Zeng, Yingsen and Liu, Yong and Wang, Hongfa and Yang, Yujiu},
  booktitle={Proceedings of the IEEE/CVF International Conference on Computer Vision},
  pages={20193--20203},
  year={2025}
}

@article{zhu2023tracking,
  title={Tracking with human-intent reasoning},
  author={Zhu, Jiawen and Cheng, Zhi-Qi and He, Jun-Yan and Li, Chenyang and Luo, Bin and Lu, Huchuan and Geng, Yifeng and Xie, Xuansong},
  journal={arXiv preprint arXiv:2312.17448},
  year={2023}
}
\bibliographystyle{plainnat}

\newpage
\appendix

\section{More implement details}

\subsection{Datasets construction}
\begin{figure}
    \centering
    \resizebox{0.9\linewidth}{!}{
        \includegraphics{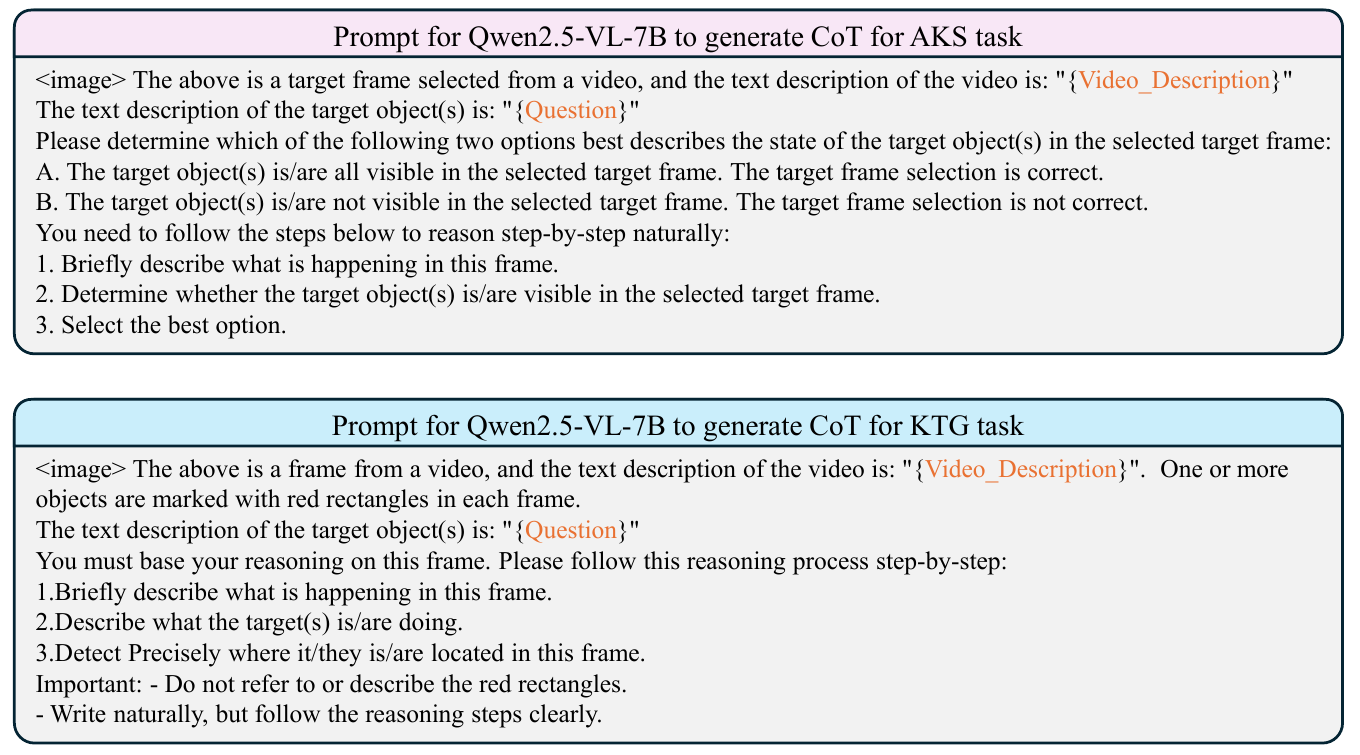}
    }
    \caption{Prompt templates for CoT datasets generation.}
    \label{fig:prompt-for-dataset}
\end{figure}

\begin{figure}
    \centering
    \resizebox{0.9\linewidth}{!}{
        \includegraphics{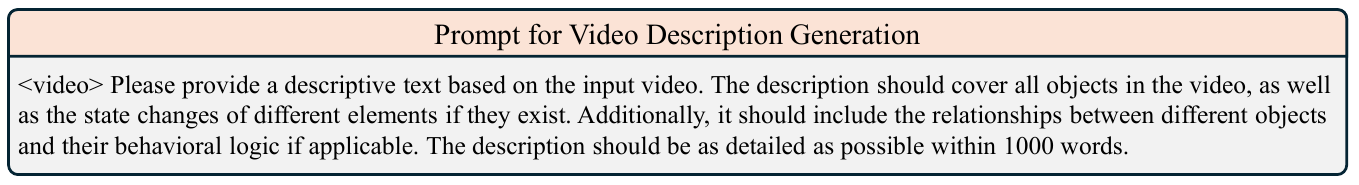}
    }
    \caption{Prompt templates for VDG task.}
    \label{fig:prompt-for-vdg}
\end{figure}

\begin{figure*}
    \centering
    \resizebox{0.9\linewidth}{!}{
        \includegraphics{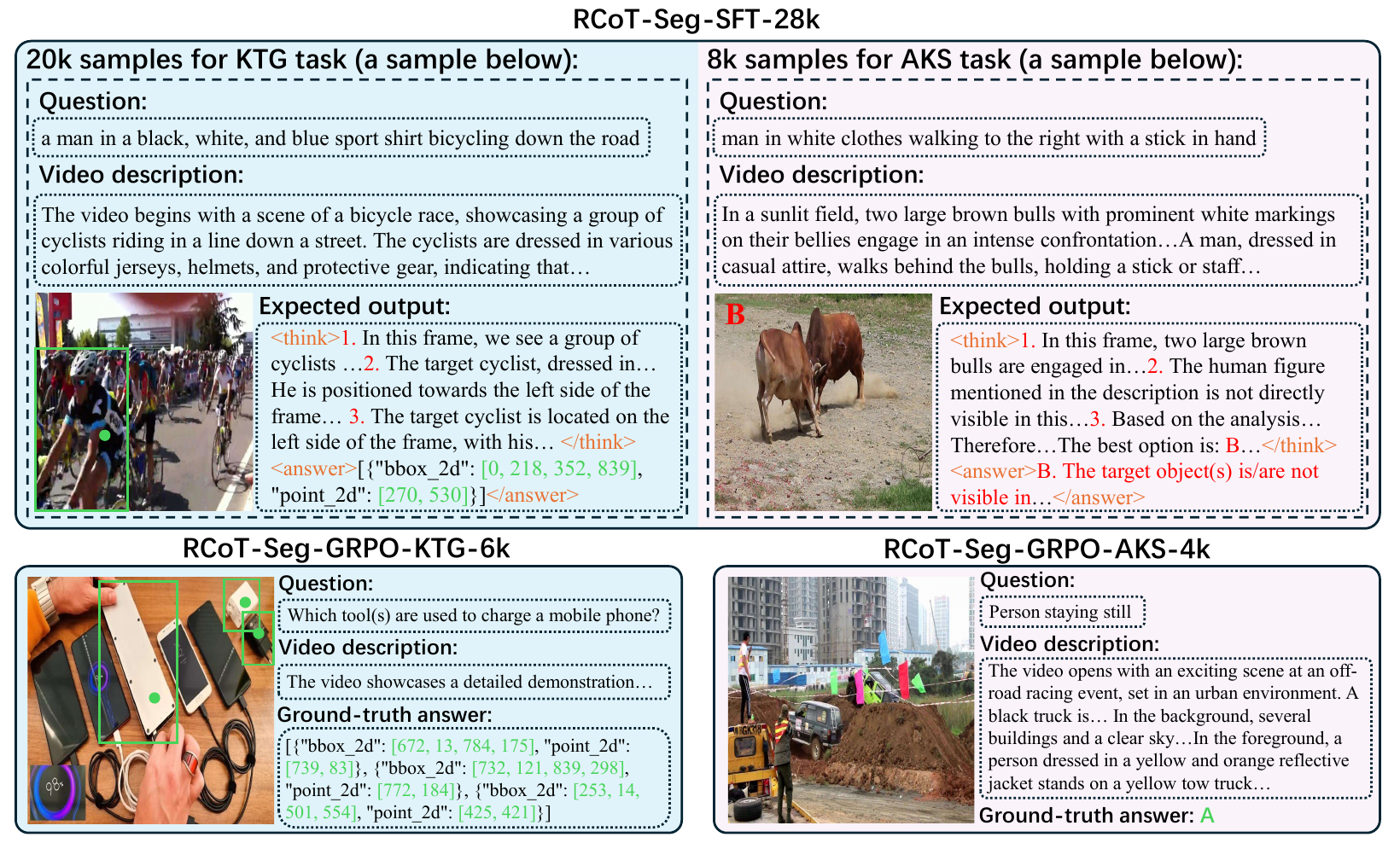}
    }
    \caption{Composition of the training datasets and sample examples.}
    \label{fig:dataset-sample}
\end{figure*}

\cref{fig:prompt-for-dataset} illustrates the prompt templates for Qwen2.5-VL-7B~\cite{bai2025qwen2} to generate Chain-of-Thought data.
Each prompt is presented to the model along with a selected keyframe (resized to $840 \times 840$) and its corresponding video description, as well as the predefined description of the target object(s).
The methods for selecting keyframes for both tasks have been discussed in the main document.
The corresponding video descriptions are generated by our base model Qwen2.5-VL-3B~\cite{bai2025qwen2} prompted by \cref{fig:prompt-for-vdg} and incorporated into our RCoT-Seg-SFT-28k, RCoT-Seg-GRPO-AKS-4k and RCoT-Seg-GRPO-KTG-6k datasets.
The prompts guide the model to (i) briefly describe the scene, (ii) analyze the target object(s), and (iii) complete the task. 
For each piece of generated CoT data, we select those that meet our task requirements and discard the lower-quality ones.
The composition of the training datasets and several sample examples are shown in \cref{fig:dataset-sample}.


\subsection{Training settings}
The training process of our model is divided into two main stages: CoT-SFT and GRPO, and the details of our training approach are shown in \cref{tab:stages}. 
In the first stage, we construct a compact Chain-of-Thought dataset RCoT-Seg-SFT-28k, and perform cold-start supervised fine-tuning (CoT-SFT) on Qwen2.5-VL-3B~\cite{bai2025qwen2} with LoRA~\cite{hu2022lora} to mitigate overfitting.
In the second stage, we use GRPO~\cite{shao2024deepseekmath,guo2025deepseek} for reinforcement learning refinement with 4K and 6K task-specific datasets, respectively.
The training is conducted on 4 NVIDIA L20 GPUs, which takes about 3 days to finish the whole multi-stage training.

During GRPO training, we design a task-aligned verifiable reward $\mathcal{R}$ to optimize model behavior, which is obtained by adding the format reward $\mathcal{R}_f$ and the answer accuracy reward $\mathcal{R}_a$ with corresponding weights $\lambda_f$ and $\lambda_a$.
For the AKS task, $\lambda_f$ and $\lambda_a$ are set to 1.0 and 2.0, respectively. The format reward checks the overall format of the output, with a full score of 1.0. The accuracy reward checks the correctness of the predicted answer, with a full score of 1.0. Therefore, the maximum total reward for the AKS task is 3.0.
For the KTG task, $\lambda_f$ and $\lambda_a$ are set to 1.0 and 1.0, respectively. The format reward further examines the format of the predicted bounding boxes and points, with a maximum score of 3.0. The accuracy reward evaluates the IoU metric and L1 distance between predicted bounding boxes and the ground truth, as well as the L1 distance between predicted points and the ground truth, with a maximum score of 3.0. The threshold hyperparameters $\eta$, $\gamma_{box}$, and $\gamma_{pt}$ are set to 0.5, 10, and 30, respectively. The maximum total reward for the KTG task is 6.0.

\begin{table*}
\rowcolors{2}{gray!25}{white}
\centering
\caption{\textbf{Training parameters.} 
}
\label{tab:stages}
\setlength{\tabcolsep}{2mm}{
\resizebox{0.9\linewidth}{!}{
\begin{tabular}{c|c|c|c}
\toprule
Config &CoT-SFT& AKS-GRPO& KTG-GRPO \\
\midrule
finetuning type &LoRA& GRPO &GRPO \\
optimizer &AdamW  &AdamW &AdamW \\
learning rate &1e-4&1e-6 &1e-6\\
weight decay &0.0 &0.01  &0.01\\
batch size &8 &16  &16 \\
lora rank &8 &- &-\\
gradient-accumulation-steps &8 &- &-\\
lr scheduler &cosine &constant &constant\\
warmup ratio &0.1&0.0&0.0\\
kl loss coef &- &0.01 &0.01 \\
kl loss type &- &low-var-kl &low-var-kl \\
rollout-n &1&8 &8 \\
temperature &0.95 &1.0 &1.0 \\
top-k & 50 &-1 &-1 \\
top-p & 0.7 & 1.0 & 1.0\\
max prompt length &3000 &3000 &3000 \\
max response length &1024 &2000 &2000 \\
training epochs &1 &- &- \\
training steps &-&500 &1000 \\
training set &RCoT-Seg-SFT-28k &RCoT-Seg-GRPO-AKS-4k &RCoT-Seg-GRPO-KTG-6k\\
\bottomrule
\end{tabular}
}
}
\end{table*}
\begin{figure*}
    \centering
    \resizebox{0.9\linewidth}{!}{
        \includegraphics{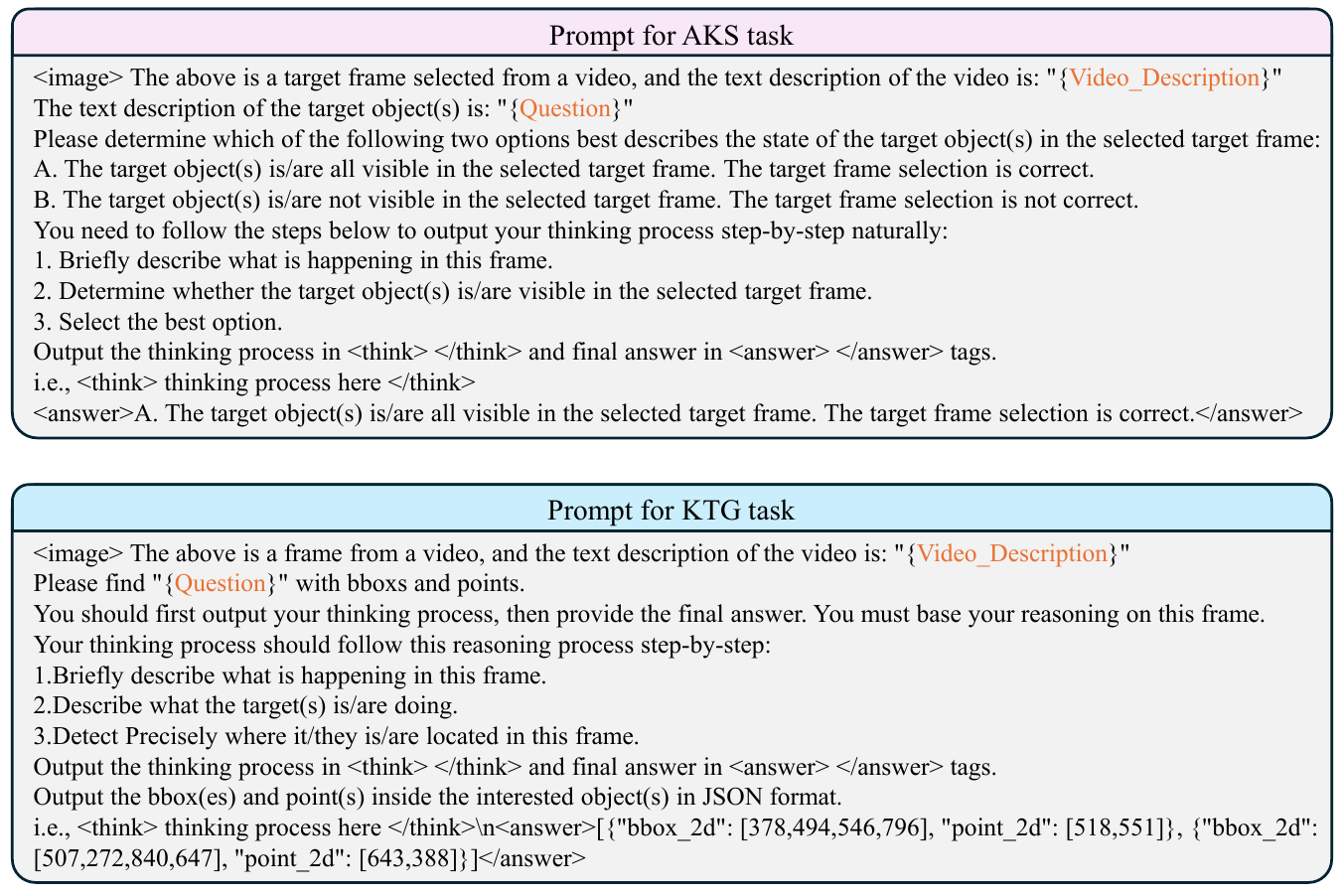}
    }
    \caption{Prompt templates for AKS and KTG tasks.}
    \label{fig:prompt-for-task}
\end{figure*}
\begin{figure*}
    \centering
    \resizebox{0.9\linewidth}{!}{
        \includegraphics{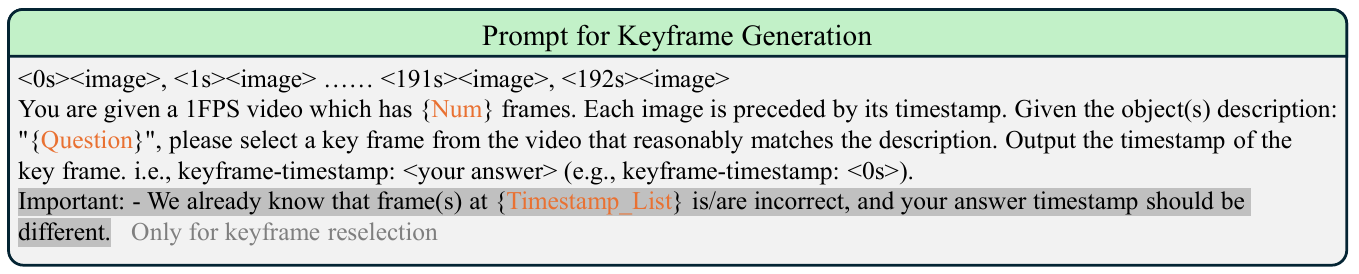}
    }
    \caption{Prompt templates for KFG task.}
    \label{fig:prompt-for-kfg}
\end{figure*}

\subsection{Inference details}
The prompt templates used for Video Description Generation and Keyframe Generation are shown in \cref{fig:prompt-for-vdg} and \cref{fig:prompt-for-kfg}, respectively. 
Additionally, the prompt templates used for AKS and KTG tasks are shown in \cref{fig:prompt-for-task}.
During Video Description Generation and Keyframe Generation, the proposed RCoT-Seg receives the full video sequence, where each frame is resized to $448 \times 448$.
In contrast, when performing AKS and KTG tasks, the input keyframe is resized to $840 \times 840$ to enhance detail perception.

\section{Effect of scaling model size}
To investigate the impact of foundation model capacity on our framework, we train a larger variant initialized from Qwen2.5-VL-7B. This model follows the exact same multi-stage training pipeline (CoT-SFT and GRPO) and hyperparameters as the 3B variant. Notably, the training datasets remain identical to those used for RCoT-Seg-3B. The video descriptions are still generated by the 3B model, and the CoT reasoning traces continue to be constructed using the 7B model.

As detailed in \cref{tab:vrs-7b} and \cref{tab:rvos-7b}, scaling the model size yields consistent improvements across VRS and RVOS benchmarks.

\begin{table*}
    \centering
    \caption{\textbf{Performance comparison on VRS datasets.} The best and second-best results are shown in \textcolor{red}{\textbf{red}} and \textcolor{blue}{\textbf{blue}} color. 
    }
    \vspace{-2mm}
    \label{tab:vrs-7b}
    \rowcolors{2}{gray!25}{white}
    \resizebox{0.98\linewidth}{!}{
        \begin{tabular}{l|ccc|ccc|ccc|ccc}
    \toprule

    \multirow{2}{*}{\textbf{Method}} & \multicolumn{3}{c|}{\textbf{ReVOS / referring}} & \multicolumn{3}{c|}{\textbf{ReVOS / reasoning}} & \multicolumn{3}{c|}{\textbf{ReVOS / overall}} & \multicolumn{3}{c}{\textbf{ReasonVOS}} \\
    \cmidrule(lr){2-4} \cmidrule(lr){4-7} \cmidrule(lr){8-10} \cmidrule(lr){11-13}
  & $\mathcal{J}\&\mathcal{F}$ & $\mathcal{J}$ & $\mathcal{F}$ & $\mathcal{J}\&\mathcal{F}$ & $\mathcal{J}$ & $\mathcal{F}$ & $\mathcal{J}\&\mathcal{F}$ & $\mathcal{J}$ & $\mathcal{F}$ & $\mathcal{J}\&\mathcal{F}$ & $\mathcal{J}$ & $\mathcal{F}$ \\
    \midrule
    \multicolumn{13}{c}{\textit{Previous models using <seg> tokens for mask prediction via SFT training}}\\
    \midrule

    VISA-7B~\cite{yan2024visa} & 50.9 &49.2 &52.6 & 43.0 &40.6 &45.4 & 46.9 &44.9 &49.0 &  -& -&-\\
    VISA-13B~\cite{yan2024visa} & 57.4 &55.6 &59.1 & 44.3 &42.0 &46.7 & 50.9 &48.8 &52.9&  -&- &-\\
    VideoLISA-3.8B~\cite{videolisa} & - &- &- &  -& -&- &  -&- &- &  47.5 &45.1 &49.9\\
    InstructSeg-3B~\cite{wei2025instructseg} & 57.0 &54.8 &59.2 & 51.9 &49.2 &54.7 & 54.5 &52.0 &56.9& - & -&-\\
    HyperSeg-3B~\cite{hyperseg} & 58.5 &56.0 &60.9 & 53.0 &50.2 &55.8 & 55.7 &53.1 &58.4&  -&- &-\\
    GLUS-7B~\cite{lin2025glus} & 58.3 &56.0 &60.7 & 51.4 &48.8 &53.9 & 54.8 &52.4 &57.3& 49.9 &47.5 &52.4\\
    VRS-HQ-7B~\cite{VRS-HQ} & 62.1 &59.8 &64.5 & 56.1 &53.5 &58.7 & 59.1 &56.6 &61.6 & -& -&-\\
    VRS-HQ-13B~\cite{VRS-HQ} & {{63.3}} &{{61.1}} &65.5 & {{56.8}} &{{54.1}} &59.4 & {{60.0}} &{{57.6}} &62.5 &  -& -&-\\

    \midrule
    \multicolumn{13}{c}{\textit{Latest models via GRPO training}}\\
    \midrule
    Omni-R1-8B~\cite{zhong2025omni} &  61.6 &- &- & 50.7 &- &- & 56.2 &- &- & - &- &- \\
    Veason-R1-3B~\cite{gong2025reinforcing} & 63.0 &60.3 &{{65.6}} & {{56.8}} &53.6 &{{60.0}} & 59.9 &56.9 &{{62.8}}& {{55.2}} &{{51.8}} &{{58.5}}\\

    \textbf{Ours-3B} 
    & \textcolor{blue}{\textbf{64.3}}~\textbf{(+1.0)}
    & \textcolor{blue}{\textbf{61.6}}
    & \textcolor{blue}{\textbf{66.9}}
    & \textcolor{blue}{\textbf{58.2}}~\textbf{(+1.4)}
    & \textcolor{blue}{\textbf{54.9}}
    & \textcolor{blue}{\textbf{61.6}}
    & \textcolor{blue}{\textbf{61.2}}~\textbf{(+1.2)}
    & \textcolor{blue}{\textbf{58.2}}
    & \textcolor{blue}{\textbf{64.3}}
    & \textcolor{blue}{\textbf{58.2}}~\textbf{(+3.0)}
    & \textcolor{blue}{\textbf{54.6}}
    & \textcolor{blue}{\textbf{61.8}}\\

    \textbf{Ours-7B} 
    & \textcolor{red}{\textbf{65.9}}~\textbf{(+2.6)}
    & \textcolor{red}{\textbf{63.2}}
    & \textcolor{red}{\textbf{68.7}}
    & \textcolor{red}{\textbf{59.7}}~\textbf{(+2.9)}
    & \textcolor{red}{\textbf{56.6}}
    & \textcolor{red}{\textbf{62.8}}
    & \textcolor{red}{\textbf{62.8}}~\textbf{(+2.8)}
    & \textcolor{red}{\textbf{59.9}}
    & \textcolor{red}{\textbf{65.7}}
    & \textcolor{red}{\textbf{59.4}}~\textbf{(+4.2)}
    & \textcolor{red}{\textbf{56.0}}
    & \textcolor{red}{\textbf{62.8}}\\
    
    \bottomrule
  \end{tabular}}
  \vspace{-4mm}
\end{table*}

\begin{table*}
    \centering
    \caption{\textbf{Performance comparison on RVOS datasets.} The best and second-best results are shown in \textcolor{red}{\textbf{red}} and \textcolor{blue}{\textbf{blue}} color.}
    \vspace{-2mm}
    \footnotesize
    \label{tab:rvos-7b}
    \rowcolors{2}{gray!25}{white}
    \resizebox{0.95\linewidth}{!}{
        \begin{tabular}{l|c|ccc|ccc|ccc}
    \toprule
    
    \rowcolors{2}{gray!25}{white}
    \multirow{2}{*}{\textbf{Method}} & \multirow{2}{*}{\textbf{MLLM}} & \multicolumn{3}{c|}{\textbf{DAVIS17}} & \multicolumn{3}{c|}{\textbf{Ref-YouTube-VOS}} & \multicolumn{3}{c}{\textbf{MeViS}}\\
    \cmidrule(lr){3-5} \cmidrule(lr){6-8} \cmidrule(lr){9-11}
 & & $\mathcal{J}\&\mathcal{F}$ & $\mathcal{J}$ & $\mathcal{F}$ & $\mathcal{J}\&\mathcal{F}$ & $\mathcal{J}$ & $\mathcal{F}$ & $\mathcal{J}\&\mathcal{F}$ & $\mathcal{J}$ & $\mathcal{F}$ \\
    \midrule

    Track-GPT-7B~\cite{zhu2023tracking} &LLaVA-7B & 63.2 &59.4 &67.0 & 56.4 &55.3 &57.4 & 40.1 &37.6 &42.6\\
    Track-GPT-13B~\cite{zhu2023tracking} &LLaVA-13B & 66.5 &62.7 &70.4 & 59.5 &58.1 &60.8 & 41.2 &39.2 &43.1\\
    VISA-7B~\cite{yan2024visa} & Chat-UniVi-7B & 69.4 &66.3 &72.5 & 61.5 &59.8 &63.2 & 43.5 &40.7 &46.3\\
    VISA-13B~\cite{yan2024visa} & Chat-UniVi-13B & 70.4 &67.0 &73.8 & 63.0 &61.4 &64.7 & 44.5 &41.8 &47.1\\
    VideoLISA-3.8B~\cite{videolisa}& LLaVA-Phi-3-V & 68.8 &59.4 &64.9 & 63.7 &61.7 &65.7 & 44.4 &41.3 &47.6\\
    GLUS-7B~\cite{lin2025glus}& LLaVA-7B & - &- &- & 67.3 &65.5 &69.0 & {{51.3}} &{{48.5}} &{{54.2}}\\
    VRS-HQ-13B~\cite{VRS-HQ} & Chat-UniVi-13B & {{74.4}} &{{71.0}} &{{77.9}} & {{71.0}} &{{69.0}} &{{73.1}} & 50.9 &48.0 &53.7\\

    \midrule
    Veason-R1-3B~\cite{gong2025reinforcing} & Qwen2.5-VL-3B & - &- &- & - &- &- & 51.2 &48.2 &{{54.2}}\\

    \textbf{Ours-3B}& Qwen2.5-VL-3B 
& \textcolor{blue}{\textbf{76.1}}~\textbf{(+1.7)}
& \textcolor{blue}{\textbf{72.2}}
& \textcolor{blue}{\textbf{80.0}}
& \textcolor{blue}{\textbf{72.5}}~\textbf{(+1.5)}
& \textcolor{blue}{\textbf{70.5}}
& \textcolor{blue}{\textbf{74.6}}
& \textcolor{blue}{\textbf{53.7}}~\textbf{(+2.4)}
& \textcolor{blue}{\textbf{50.7}}
& \textcolor{blue}{\textbf{56.7}}\\

    \textbf{Ours-7B}& Qwen2.5-VL-7B 
& \textcolor{red}{\textbf{78.6}}~\textbf{(+4.2)}
& \textcolor{red}{\textbf{75.0}}
& \textcolor{red}{\textbf{82.2}}
& \textcolor{red}{\textbf{73.3}}~\textbf{(+2.3)}
& \textcolor{red}{\textbf{71.5}}
& \textcolor{red}{\textbf{75.1}}
& \textcolor{red}{\textbf{54.8}}~\textbf{(+3.5)}
& \textcolor{red}{\textbf{51.9}}
& \textcolor{red}{\textbf{57.7}}\\
    
    \bottomrule
  \end{tabular}}
  \vspace{-4mm}
\end{table*}

\section{Timing analysis}
\label{appendix:time-analysis}
\begin{figure*}
    \centering
    \resizebox{0.65\linewidth}{!}{
        \includegraphics{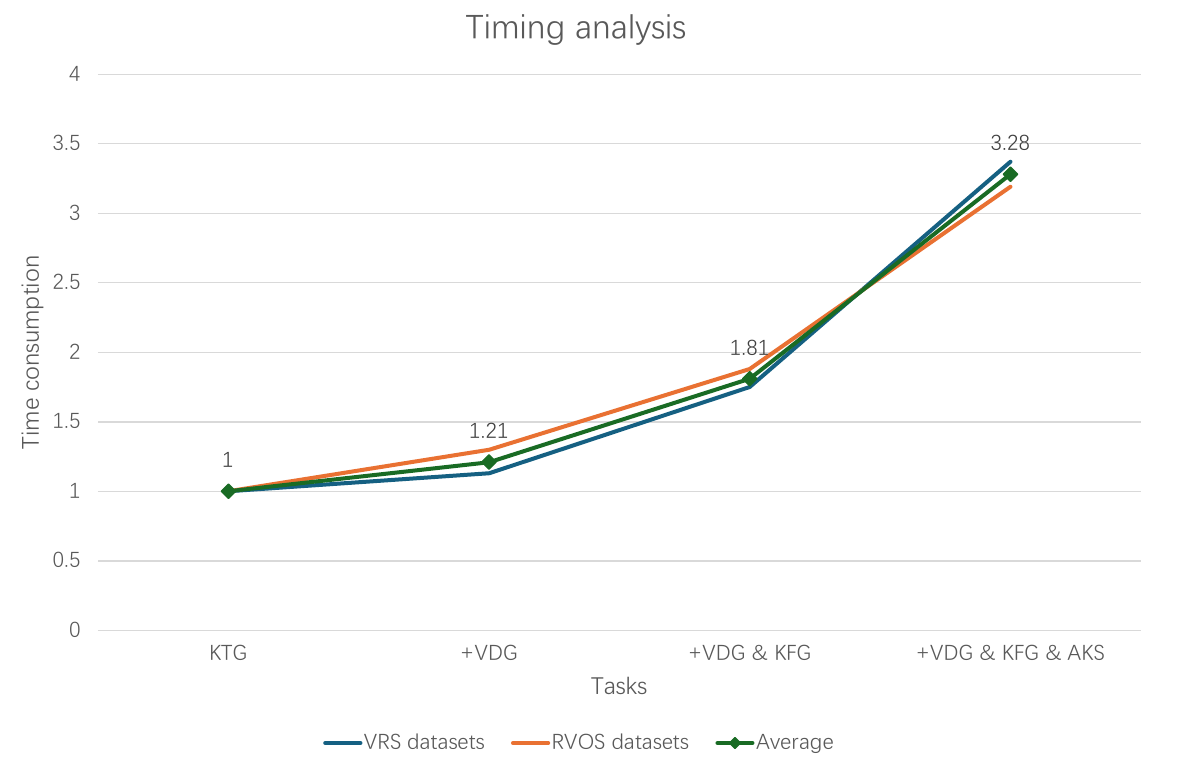}
    }
    \caption{Curves of timing analysis on VRS and RVOS datasets. }
    \label{fig:time-curves}
\end{figure*}

We analyze the additional time consumption required to complete VDG, KFG, and AKS tasks compared to performing only the KTG task. We set the time consumption for the KTG task as 1.0, which includes the time taken by SAM2~\cite{ravi2024sam} for keyframe segmentation and propagation. As shown in \cref{fig:time-curves}, completing the VDG, KFG, and AKS tasks requires additional time consumption of 0.21, 0.60, and 1.47 respectively. Without using AKS, the inference time consumption is 1.81. The additional time consumption incurred by adopting AKS is 1.47. Therefore, implementing the AKS mechanism results in an 81\% increase in additional time consumption.

Our evaluation reveals that the latency overhead primarily stems from repeated full-video encoding during the AKS re-evaluation phase, rather than from the AKS mechanism itself. To verify this and optimize efficiency, we introduce a light re-evaluation strategy: while maintaining the initial encoding pass on the full video, we replace subsequent re-evaluations with 16 uniformly sampled frames. As shown in \cref{tab:inference_time_analysis}, this sampling strategy substantially reduces the extra computational overhead from 81\% to 40\% while maintaining highly competitive performance.
\begin{table}[htbp]
  \centering
  \caption{Performance and computational overhead comparison of different AKS strategies.}
  \label{tab:inference_time_analysis}
  \resizebox{1.0\linewidth}{!}{
  \begin{tabular}{l c c c c c}
    \toprule
    \textbf{Method} & \textbf{Extra Time} & \textbf{Ref-YT-VOS} & \textbf{ReasonVOS} & \textbf{ReVOS} & \textbf{MeViS} \\
    \midrule
    Ours (Full Video)                  & 81\% & 72.5 & 58.2 & 61.2 & 53.7\\
    Ours (Sample 16 after Full Video)  & \textbf{40\%} & 72.4 & 57.8 & 60.9 & 54.2\\
    \bottomrule
  \end{tabular}}
\end{table}

Furthermore, we report the average inference time per video in \cref{tab:infer-time}, split into keyframe selection and segmentation. This breakdown shows that the main runtime bottleneck in VRS remains the segmentation stage (MLLM + SAM2), not keyframe selection. In particular, AKS adds only a small overhead relative to VRS-HQ (12s vs. 11s), while our downstream segmentation stage is actually faster than both prior methods. Thus, the agentic design does not make the overall pipeline disproportionately heavy.
\begin{table}[htbp]
\centering
\caption{Inference time analysis.}
\label{tab:infer-time}
\begin{tabular}{lcc}
\toprule
\textbf{Method} & \textbf{Keyframe Selection Time} & \textbf{Segmentation Time} \\
\midrule
VISA & 21s & 24s \\
VRS-HQ & 11s & 28s \\
\textbf{Ours} & \textbf{12s (AKS)} & \textbf{19s} \\
\bottomrule
\end{tabular}
\end{table}

We also conduct a direct efficiency comparison with our concurrent work Veason-R1 in \cref{tab:infer-time-veason}. Our evaluation reveals that Veason-R1-3B takes a total of 36 seconds per video. This latency occurs because its architecture requires feeding multiple high-resolution frames simultaneously into the MLLM. Due to severe memory constraints, it can only support a maximum of 8 sampled frames as input on a 48GB L20 GPU.
\begin{table}[htbp]
\centering
\caption{Inference time comparison with Veason-R1.}
\label{tab:infer-time-veason}
\begin{tabular}{lc}
\toprule
\textbf{Method} & \textbf{Keyframe Selection and Segmentation Time}\\
\midrule
Veason-R1-3B (Sample 8 frames) & 36s \\
\textbf{Ours} & \textbf{31s} \\
\bottomrule
\end{tabular}
\end{table}

\section{Intermediate-level evaluations}

\textbf{Keyframe quality and re-selection efficiency.} To measure keyframe quality, we compute the ratio between the target object’s pixel area in the selected frame and its maximum area over the full video. As shown in \cref{tab:keyframe-quality-and-rounds}, AKS consistently replaces suboptimal initial frames with more informative ones, substantially increasing target visibility, while requiring only 1.3–2.4 rounds on average.

\begin{table}[htbp]
\centering
\caption{Keyframe quality and re-selection efficiency.}
\label{tab:keyframe-quality-and-rounds}
\resizebox{0.95\linewidth}{!}{
\begin{tabular}{llcccc}
\toprule
\textbf{Metric} & \textbf{Setting} & \textbf{Davis} & \textbf{Ref-Youtube-VOS} & \textbf{ReasonVOS} & \textbf{ReVOS} \\
\midrule
\textbf{Area Ratio (\%)} & Initial & 43.0 & 51.6 & 37.3 & 61.0 \\
 & Reselected & \textbf{65.9 (+22.9)} & \textbf{64.1 (+12.5)} & \textbf{58.0 (+20.7)} & \textbf{67.9 (+6.9)} \\
\textbf{Average Rounds} & - & 1.4 & 1.3 & 2.3 & 2.4 \\
\bottomrule
\end{tabular}}
\end{table}

\textbf{Intermediate grounding accuracy.} To evaluate spatial precision before SAM2 propagation, we compute the Avg IoU between predicted intermediate boxes and ground-truth boxes (converted from masks). The IoU values in \cref{tab:inter-gd-acc} show that the KTG stage already provides accurate localization before mask propagation.

\begin{table}[htbp]
\centering
\caption{Intermediate grounding accuracy.}
\label{tab:inter-gd-acc}
\begin{tabular}{lcccc}
\toprule
\textbf{Metric} & \textbf{Davis} & \textbf{Ref-Youtube-VOS} & \textbf{ReasonVOS} & \textbf{ReVOS} \\
\midrule
\textbf{Avg IoU (\%)} & 75.0 & 70.4 & 57.7 & 60.2 \\
\bottomrule
\end{tabular}
\end{table}

\textbf{Necessity of Explicit CoT.} To directly evaluate the role of explicit CoT traces, we train an Answer-Only baseline by removing the <think> ... </think> content from the training targets and asking the model to predict only the final answer. The SAM2 interface, output format, and all other training settings are kept unchanged, so the comparison isolates the effect of explicit reasoning supervision.

As shown in \cref{tab:necessity-of-ecot},the gains are consistent across all benchmarks, showing that explicit CoT supervision provides a clear benefit beyond predicting the final answer alone. The improvements are especially larger on ReasonVOS (+3.1) and ReVOS (+2.9), suggesting that explicit reasoning helps build a stronger intermediate representation for keyframe verification and target grounding, rather than serving as a simple formatting choice.

\begin{table}[htbp]
\centering
\caption{Necessity of Explicit CoT.}
\label{tab:necessity-of-ecot}
\begin{tabular}{lcccc}
\toprule
\textbf{Method} & \textbf{Davis} & \textbf{Ref-Youtube-VOS} & \textbf{ReasonVOS} & \textbf{ReVOS} \\
\midrule
Answer-Only & 73.9 & 69.8 & 55.1 & 58.3 \\
\textbf{Explicit-CoT (Ours)} & \textbf{76.1(+2.2)} & \textbf{72.5(+2.7)} & \textbf{58.2(+3.1)} & \textbf{61.2(+2.9)} \\
\bottomrule
\end{tabular}
\end{table}

\begin{figure}
    \centering
    \begin{subfigure}[b]{0.32\textwidth}
        \centering
        \includegraphics[width=\textwidth]{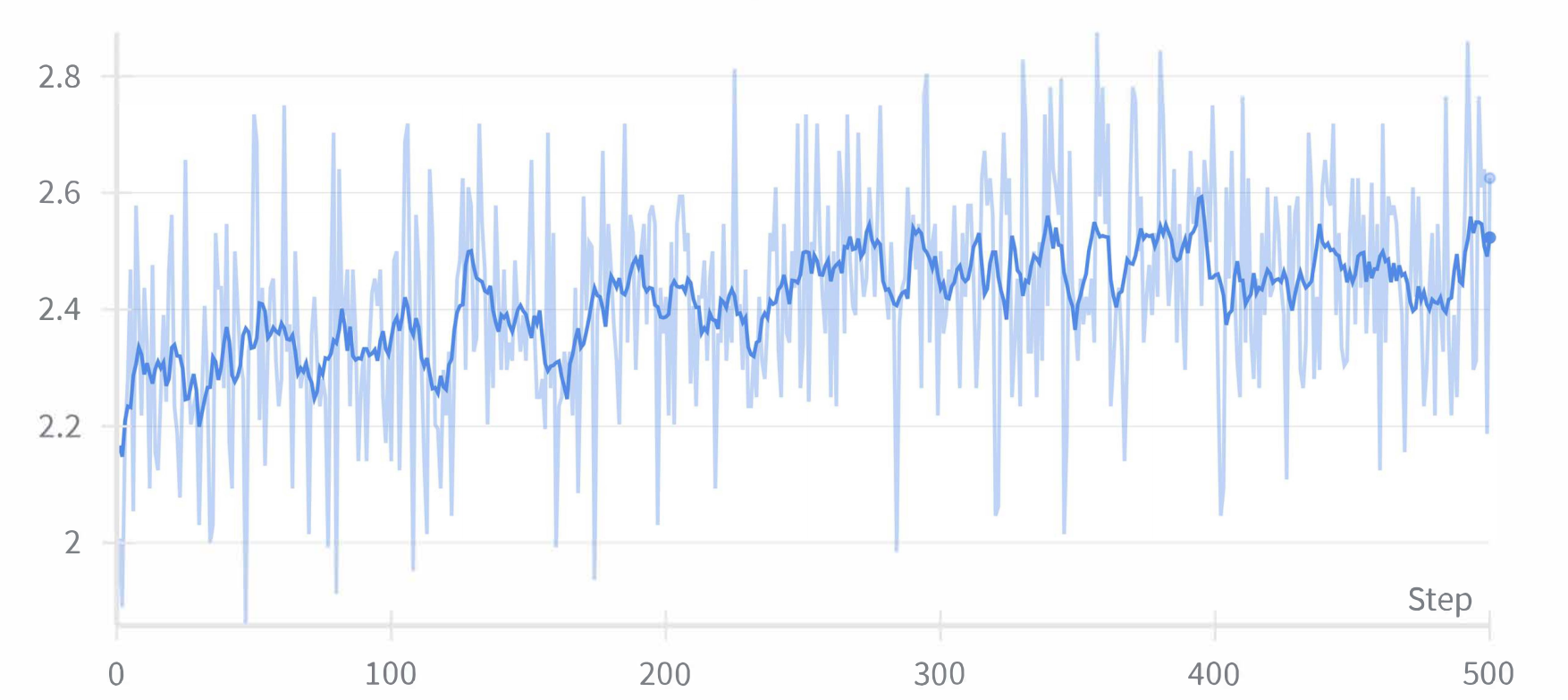}
        \caption{Average reward}
        \label{fig:1-sub1}
    \end{subfigure}
    \hfill
    \begin{subfigure}[b]{0.32\textwidth}
        \centering
        \includegraphics[width=\textwidth]{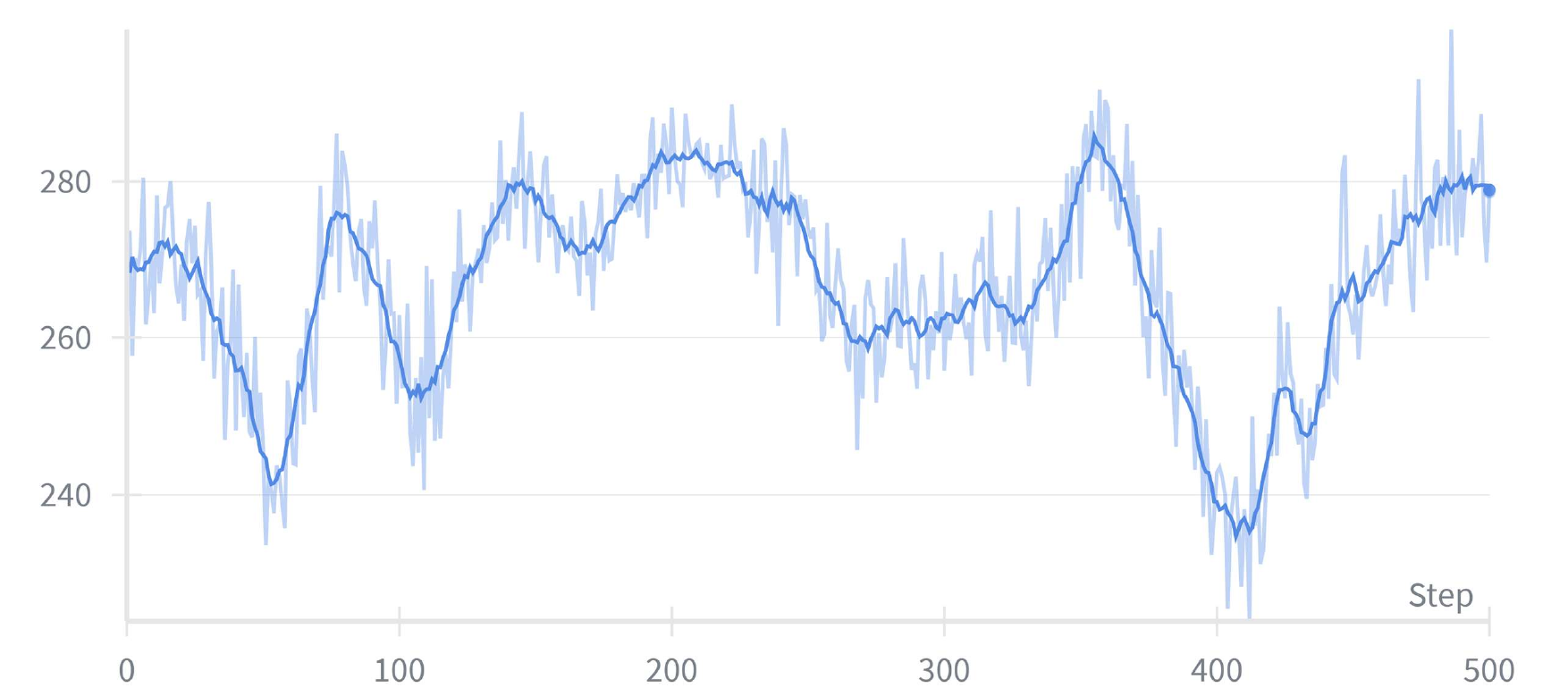}
        \caption{Average response length}
        \label{fig:1-sub2}
    \end{subfigure}
    \hfill
    \begin{subfigure}[b]{0.32\textwidth}
        \centering
        \includegraphics[width=\textwidth]{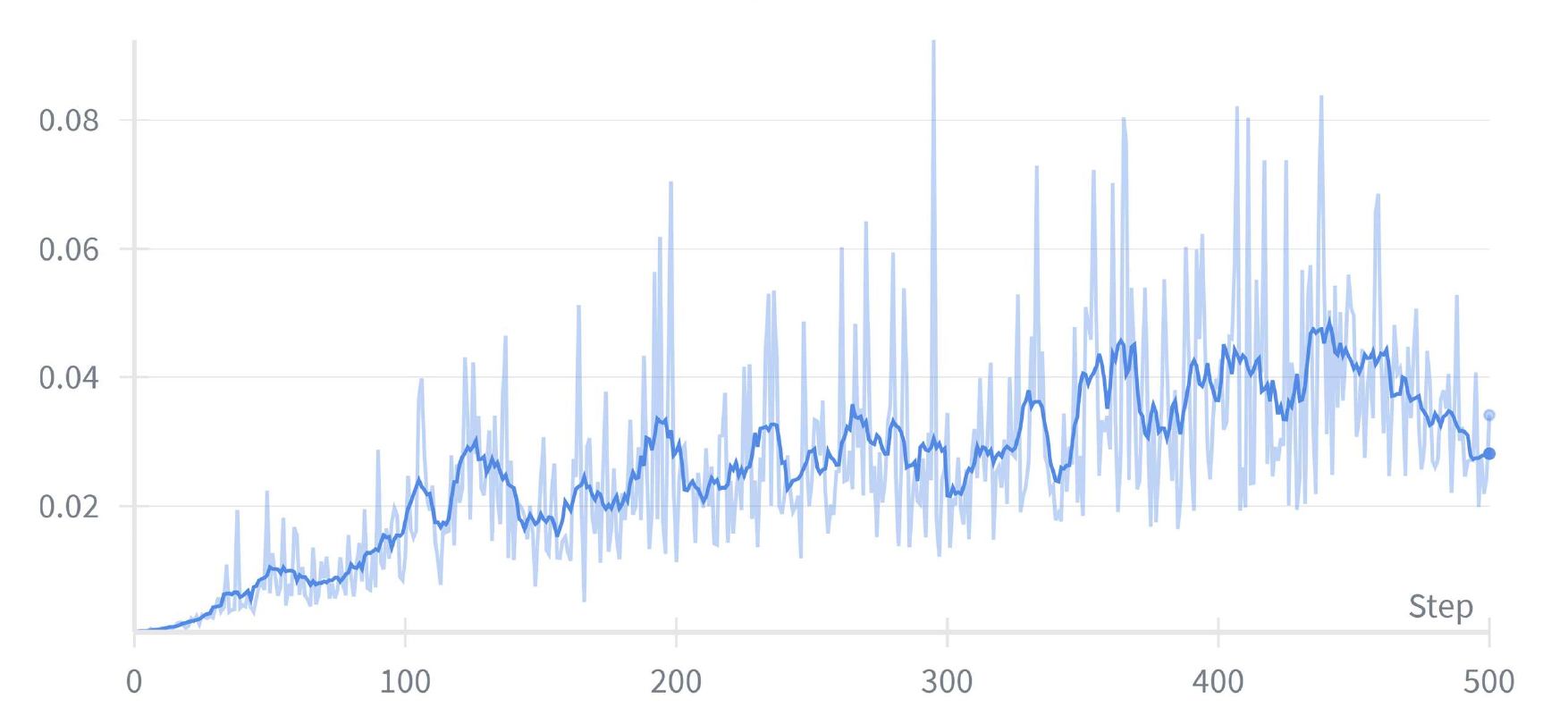}
        \caption{KL Loss}
        \label{fig:1-sub3}
    \end{subfigure}
    \caption{Curves of GRPO training for AKS task.}
    \label{fig:1-curves}
\end{figure}
\begin{figure}
    \centering
    \begin{subfigure}[b]{0.32\textwidth}
        \centering
        \includegraphics[width=\textwidth]{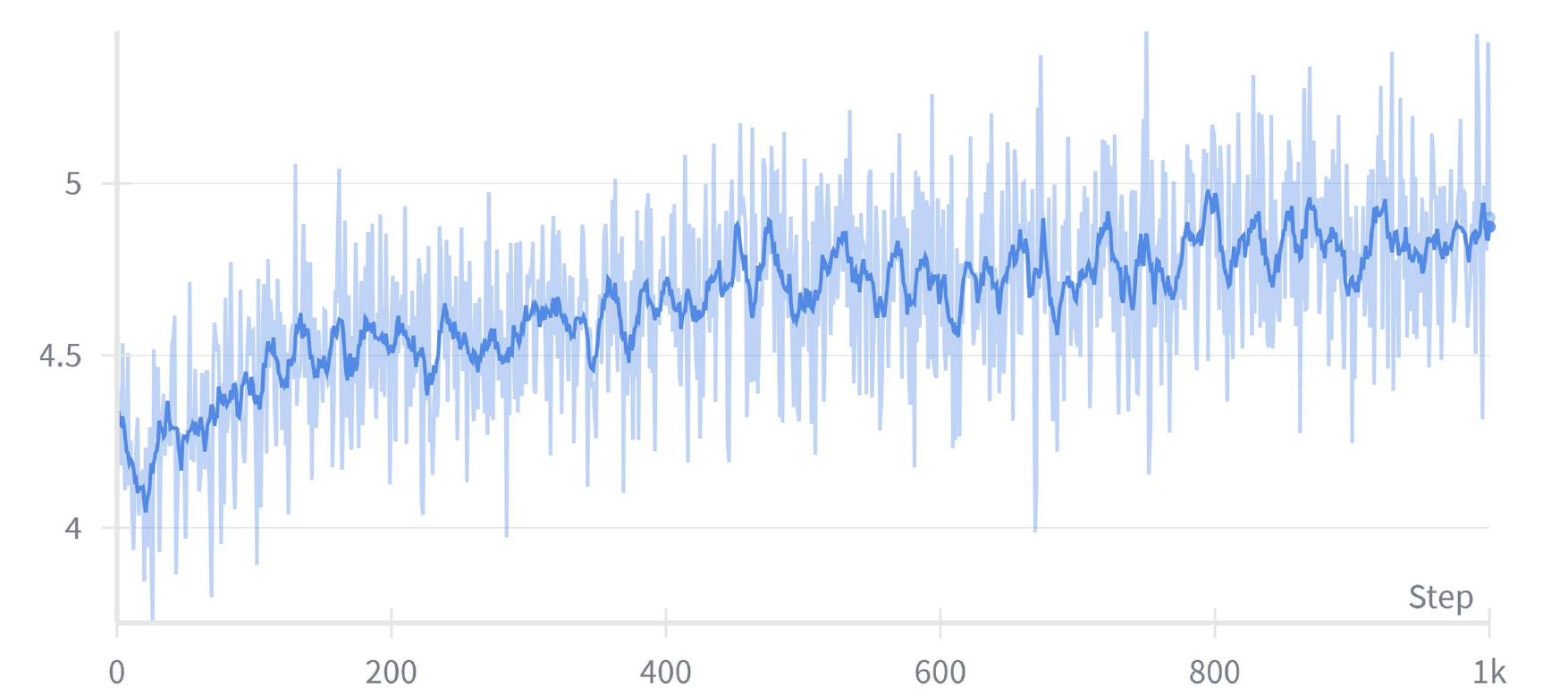}
        \caption{Average reward}
        \label{fig:2-sub1}
    \end{subfigure}
    \hfill
    \begin{subfigure}[b]{0.32\textwidth}
        \centering
        \includegraphics[width=\textwidth]{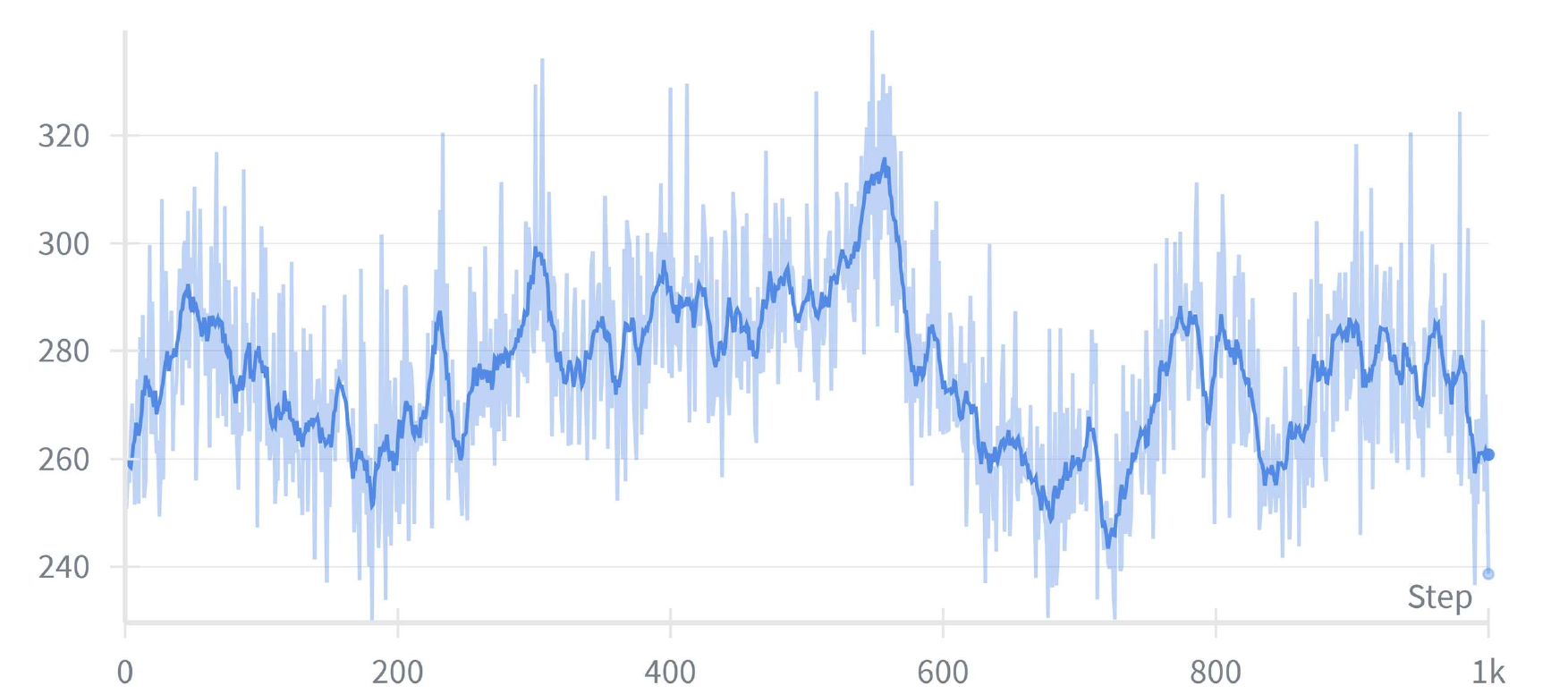}
        \caption{Average response length}
        \label{fig:2-sub2}
    \end{subfigure}
    \hfill
    \begin{subfigure}[b]{0.32\textwidth}
        \centering
        \includegraphics[width=\textwidth]{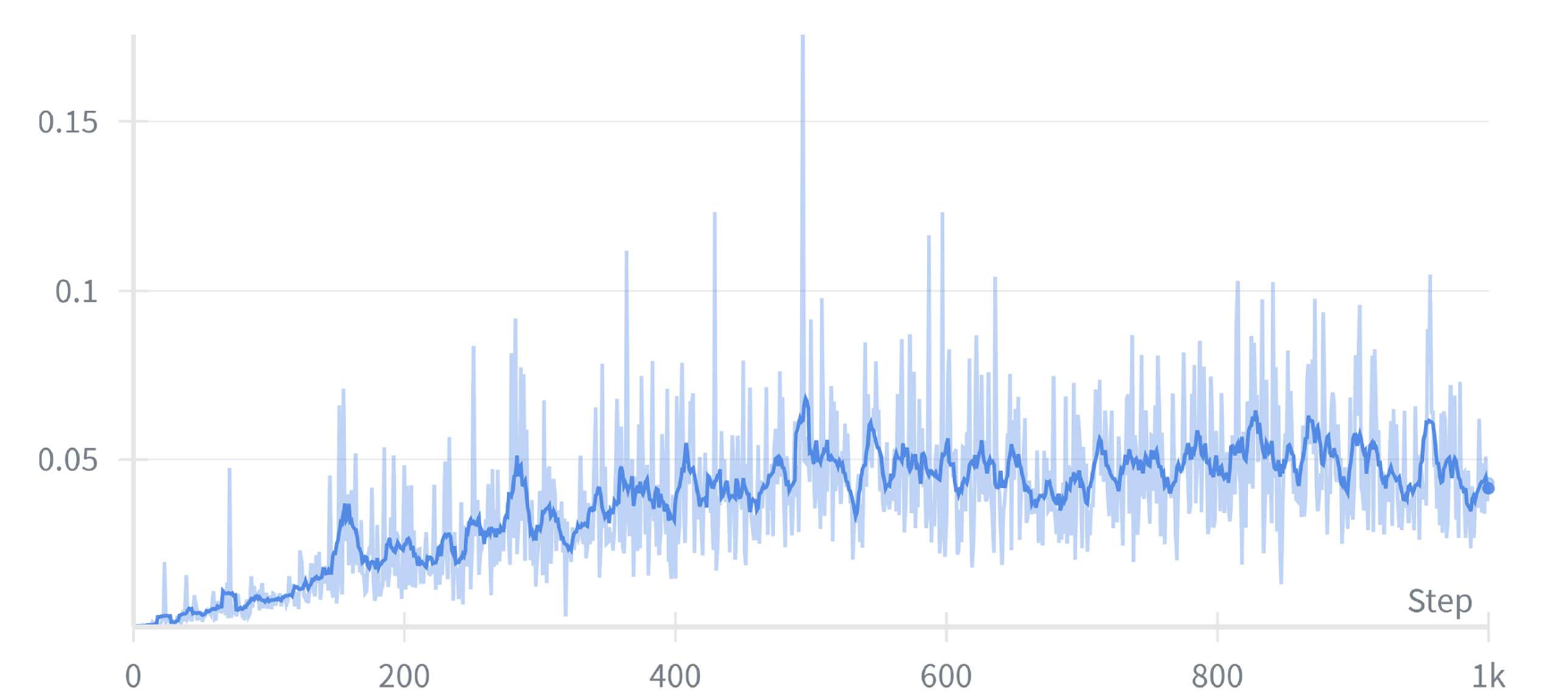}
        \caption{KL Loss}
        \label{fig:2-sub3}
    \end{subfigure}
    \caption{Curves of GRPO training for KTG task.}
    \label{fig:2-curves}
\end{figure}

\section{Training curves analysis}

\begin{figure}
    \centering
    \resizebox{0.5\linewidth}{!}{
        \includegraphics{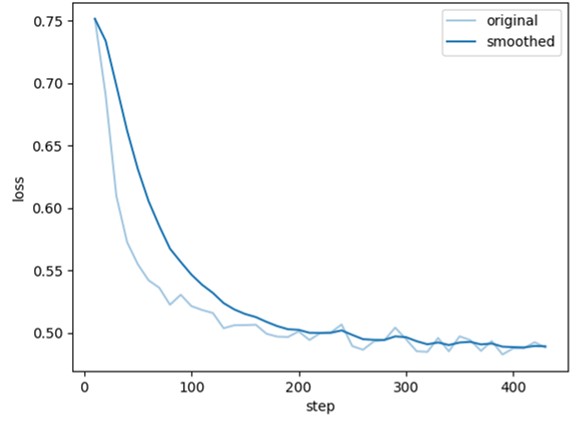}
    }
    \caption{Curve of CoT-SFT training loss. }
    \label{fig:sft-curve}
\end{figure}

We visualize the GRPO training dynamics of the proposed RCoT-Seg in \cref{fig:1-curves,fig:2-curves}, as well as the CoT-SFT training loss in \cref{fig:sft-curve}.

We monitor three key metrics during GRPO training: task-aligned reward $\mathcal{R}$, response length, and the value of KL loss. As shown in \cref{fig:1-sub1,fig:2-sub1}, the model initialized via CoT-based supervision starts with a reward score around 2.1 in AKS task and 4.3 in KTG task, indicating a reasonable ability to generate structured outputs for both tasks. The consistent upward trend in reward reflects effective policy refinement and better alignment with the designed reward function. As can be seen from \cref{fig:1-sub2,fig:2-sub2}, the average response length of our model for both tasks remains stable at around 300 tokens. As depicted in \cref{fig:1-sub3,fig:2-sub3}, the KL divergence progressively increases as the model shifts away from the initial policy to explore more reward-driven behaviors. It eventually plateaus, signifying convergence toward a consistent and reward-aligned policy under the regularization constraint. 

The CoT-SFT training loss of the model shown in \cref{fig:sft-curve} demonstrates a consistent and significant decrease from approximately 0.75 to around 0.48 over 400 steps, indicating effective learning convergence.

\section{Limitations}
In this study, the construction of the segmentation dataset does not incorporate samples for zero-target segmentation~\cite{xia2024gsva}, which limits the model from to handle zero-target segmentation tasks. When the instruction refers to an object absent from the video, the model consequently produces incorrect segmentation results,  which could be mitigated by developing a more comprehensive dataset and modifying the corresponding evaluation strategy. Furthermore, regarding dataset construction, unlike some existing methods that rely on massive frontier models for data synthesis, our approach utilizes the relatively smaller open-weight Qwen2.5-VL series. Specifically, we employ Qwen2.5-VL-7B to construct the Chain-of-Thought reasoning traces and Qwen2.5-VL-3B to generate the video descriptions.

\section{Failure case analysis}
\begin{figure*}
    \centering
    \resizebox{\linewidth}{!}{
        \includegraphics{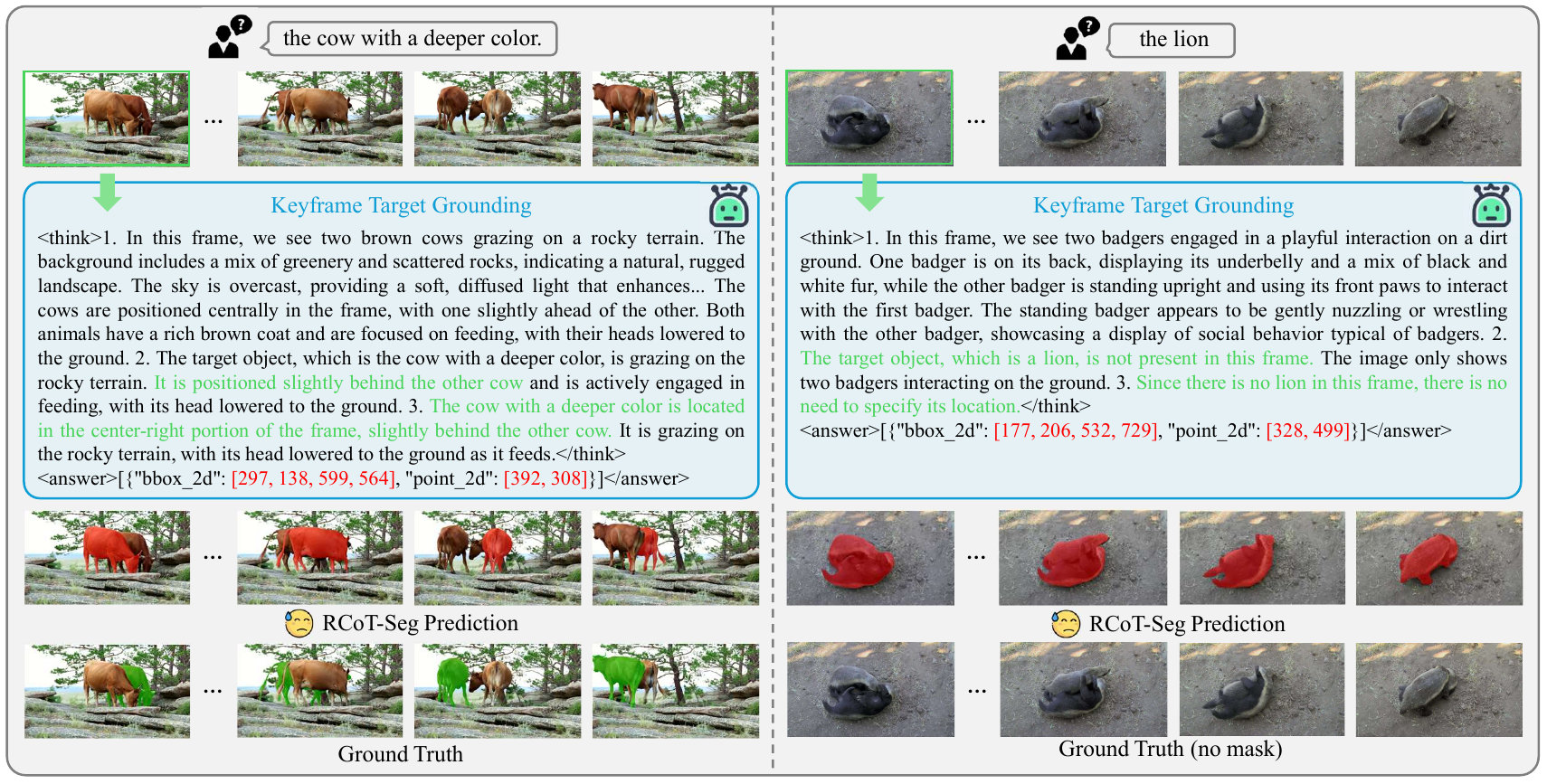}
    }
    \caption{Visualization of failure cases on the ReVOS dataset. RCoT-Seg reveals inconsistencies between its reasoning traces and final segmentation outputs, and demonstrates limited adaptability in the zero-target task.}
    \label{fig:bad-case-1}
\end{figure*}

\begin{figure*}
    \centering
    \resizebox{0.85\linewidth}{!}{
        \includegraphics{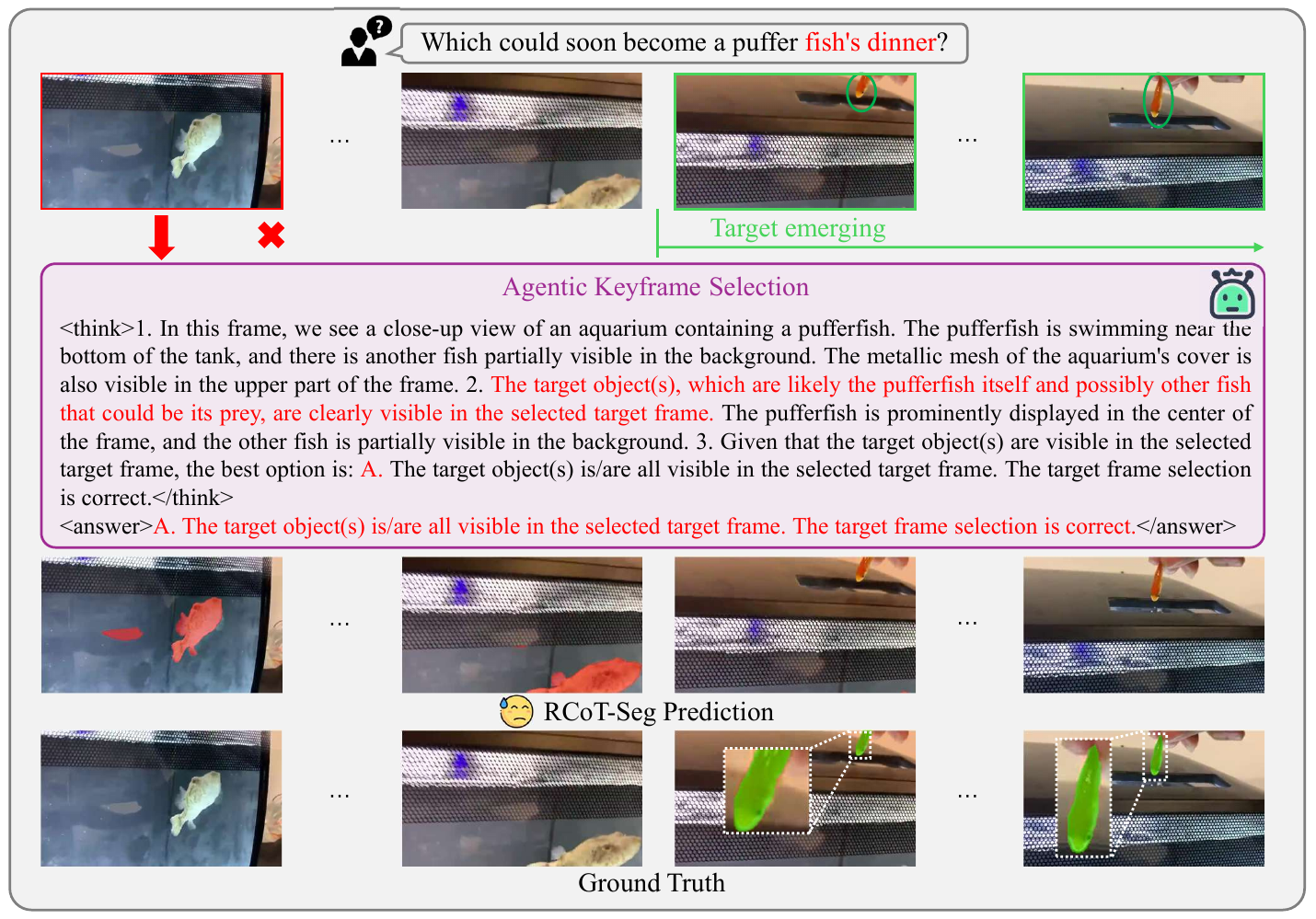}
    }
    \caption{Visualization of failure cases on the ReVOS dataset. In this case, RCoT-Seg erroneously classifies an invalid keyframe as valid in the AKS task.}
    \label{fig:bad-case-2}
\end{figure*}

\cref{fig:bad-case-1} presents some failure cases of RCoT-Seg on the ReVOS~\cite{yan2024visa} dataset. In the left example,  the model generates a correct reasoning trace, accurately identifying the cow with a deeper color and describing its relation to another cow. However, the predicted masks fail to align with the described target, revealing an inconsistency between the reasoning trace and the final segmentation outputs. In the right case, it is a zero-target scenario where the lion is not present in the video sequence. The model correctly identifies this situation in its reasoning trace, and yet it still outputs incorrect localization information, demonstrating its limited adaptability in the zero-target task. \cref{fig:bad-case-2} further reveals the misjudgment of RCoT-Seg when performing the AKS task. In this case, the target object has not yet appeared in the keyframe deemed correct by the model, resulting in an incorrect segmentation.


\section{More qualitative comparison}
We present additional visual comparisons of RCoT-Seg-3B with GLUS-7B~\cite{lin2025glus}
 on the ReasonVOS~\cite{videolisa} dataset and VRS-HQ-7B on the ReVOS~\cite{yan2024visa} dataset, to underscore its robust reasoning capabilities
 and enhanced fine-grained grounding performance.

\begin{figure*}
    \centering
    \resizebox{\linewidth}{!}{
        \includegraphics{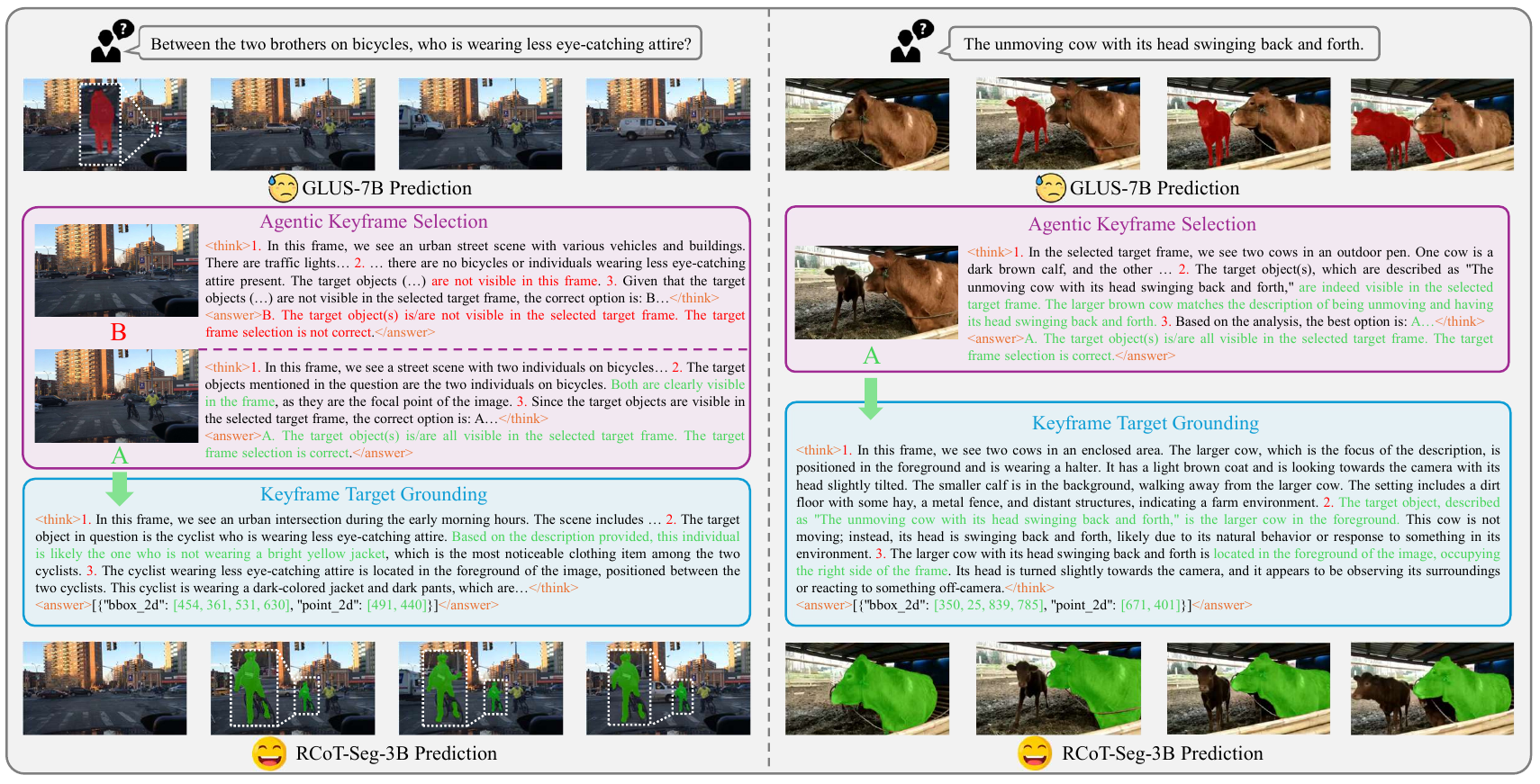}
    }
    \caption{More qualitative comparison on the ReasonVOS dataset.}
    \label{fig:vis-reasonvos}
\end{figure*}

The qualitative comparison of RCoT-Seg-3B with GLUS-7B is shown in \cref{fig:vis-reasonvos}. In the left example, various pedestrians, vehicles, and traffic facilities at the intersection constitute a complex video scene. GLUS-7B produces an incorrect segmentation result. In contrast, the proposed RCoT-Seg-3B first accurately identifies the incorrectly selected keyframe through the AKS task and reselects the correct keyframe. Then, in the KTG task, the model correctly identifies the target object (who is not wearing a bright yellow jacket) based on the provided video description and accurately predicts the segmentation masks. In the right case, the query involves action logic (unmoving, with its head swinging back and forth), which requires the model to perform reasoning. RCoT-Seg-3B accurately identifies the correct target object (the larger cow in the foreground) in the reasoning trace and provides an accurate masks prediction, whereas GLUS-7B produces an incorrect answer.

\begin{figure*}
    \centering
    \resizebox{\linewidth}{!}{
        \includegraphics{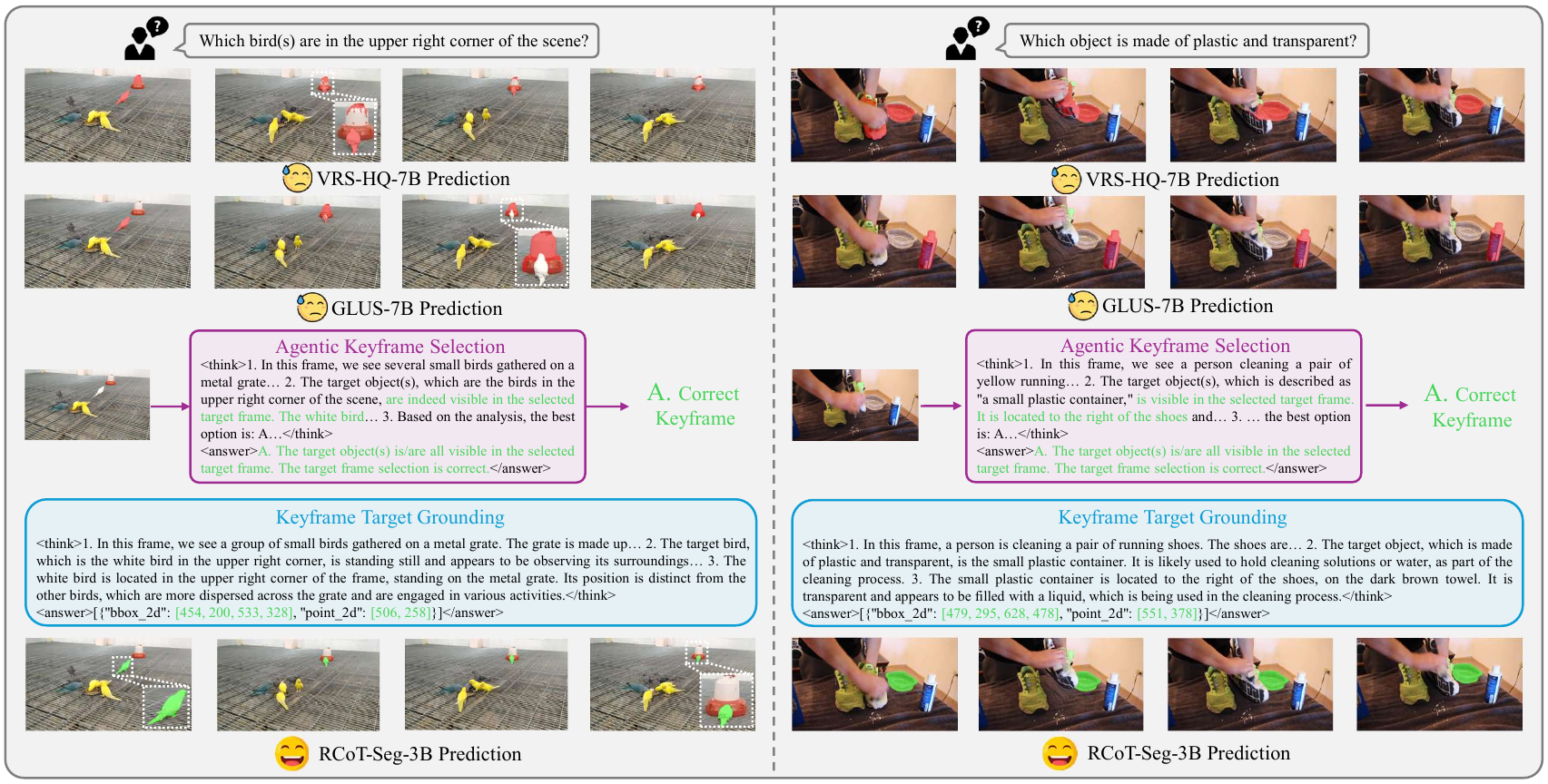}
    }
    \caption{More qualitative comparison on the ReVOS dataset.}
    \label{fig:vis-revos}
\end{figure*}

\cref{fig:vis-revos} visualizes the segmentation maps of RCoT-Seg-3B, GLUS-7B~\cite{lin2025glus} and VRS-HQ-7B~\cite{VRS-HQ} on the ReVOS dataset. In the left example, RCoT-Seg-3B correctly segments the bird in the upper right corner of the scene, whereas both VRS-HQ-7B and GLUS-7B confuse this bird with another incorrect target object due to their overlap. In the right case, RCoT-Seg-3B accurately segments the target object (transparent, made of plastic), whereas both GLUS-7B and VRS-HQ-7B confuse it with other objects, demonstrating our model's superior ability to comprehend object materials and colors.





\end{document}